\documentclass[times,twocolumn,final]{elsarticle}

\usepackage{medima}
\usepackage{framed,multirow}

\usepackage{amssymb}
\usepackage{latexsym}

\usepackage{url}
\usepackage{color,colortbl,booktabs}
\usepackage[table,x11names,dvipsnames,table]{xcolor}
\definecolor{lightgray}{gray}{0.95}

\usepackage{hyperref}
\hypersetup{
    colorlinks=true,
    linkcolor=blue,
}

\usepackage{float,tablefootnote,threeparttable,wrapfig}

\usepackage{booktabs}
\usepackage{subcaption}

\usepackage[multiple]{footmisc}

\usepackage{lipsum}
\usepackage{color}
\usepackage[ruled,vlined]{algorithm2e}
\usepackage{diagbox} 
\usepackage{tabu}
\usepackage{arydshln} 

\usepackage{makecell}

\newcommand{\revision}[1]{\textcolor{black}{#1}} 
\newcommand{\secRevision}[1]{\textcolor{black}{#1}} 
\newcommand{\final}[1]{\textcolor{black}{#1}} 

\journal{Medical Image Analysis}

\begin{document}

\verso{Z. Jiang \textit{et~al.}}

\begin{frontmatter}

\title{Robotic Ultrasound Imaging: State-of-the-Art and Future Perspectives}

\author[1]{Zhongliang \snm{Jiang}\corref{cor1}}
\cortext[cor1]{Corresponding author at: Technische Universit{\"a}t M{\"u}nchen, Fakult{\"a}t f{\"u}r Informatik -- I16, Boltzmannstr. 3, 85748 Garching bei M{\"u}nchen}
\ead{zl.jiang@tum.de}

\author[2]{Septimiu E. \snm{Salcudean}}

\author[1,3]{Nassir \snm{Navab}}

\address[1]{Computer Aided Medical Procedures, Technical University of Munich, Munich, Germany}
\address[2]{Department of Electrical and Computer Engineering, University of British Columbia, Vancouver, BC V6T 1Z4, Canada}
\address[3]{Computer Aided Medical Procedures, Johns Hopkins University, Baltimore, MD, USA}

\received{XX June 2021}
\finalform{xx Month 2021}
\accepted{xx Month 2021}
\availableonline{xx Month 2021}

\begin{abstract}
Ultrasound (US) is one of the most widely used modalities for clinical intervention and diagnosis due to the merits of providing non-invasive, radiation-free, and real-time images. However, free-hand US examinations are highly operator-dependent. Robotic US System (RUSS) aims at overcoming this shortcoming by offering reproducibility, while also aiming at improving dexterity, and intelligent anatomy and disease-aware imaging. In addition to enhancing diagnostic outcomes, RUSS also holds the potential to provide medical interventions for populations suffering from the shortage of experienced sonographers. In this paper, we categorize RUSS as teleoperated or autonomous. Regarding teleoperated RUSS, we summarize their technical developments, and clinical evaluations, respectively. This survey then focuses on the review of recent work on autonomous robotic US imaging. We demonstrate that machine learning and artificial intelligence present the key techniques, which enable intelligent patient and process-specific, motion and deformation-aware robotic image acquisition.
We also show that the research on artificial intelligence for autonomous RUSS has directed the research community toward understanding and modeling expert sonographers' semantic reasoning and action. Here, we call this process, the recovery of the ``language of sonography". This side result of research on autonomous robotic US acquisitions could be considered as valuable and essential as the progress made in the robotic US examination itself. 
This article will provide both engineers and clinicians with a comprehensive understanding of RUSS by surveying underlying techniques. 
Additionally, we present the challenges that the scientific community needs to face in the coming years in order to achieve its ultimate goal of developing intelligent robotic sonographer colleagues. These colleagues are expected to be capable of collaborating with human sonographers in dynamic environments to enhance both diagnostic and intraoperative imaging. 
\end{abstract}


\end{frontmatter}


\section{Introduction}
\label{introduction}

\begin{figure}[ht!]
\centering
\includegraphics[width=0.45\textwidth]{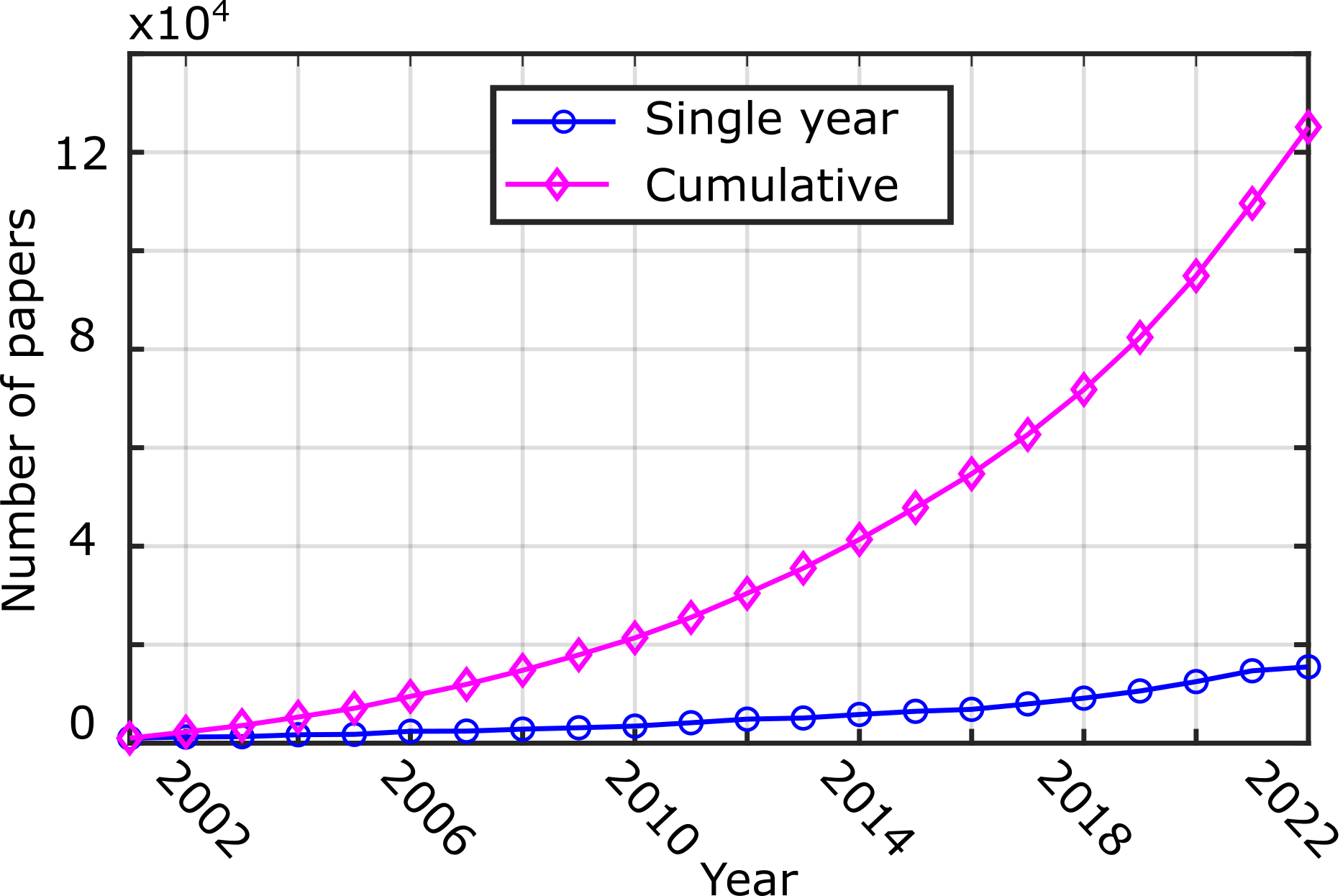}
\caption{The Number of \secRevision{peer-reviewed articles} related to RUSS. Data were collected from Google Scholar in \revision{February 2023} using the keywords: \revision{``robotic or robot" and ``ultrasound or ultrasonography" and ``Imaging or screening or scanning or acquisition".}
}
\label{Fig_publication_numbers}
\end{figure}

Today, medical imaging is one of the most crucial components of the entire healthcare industry, from wellness and screening to early diagnosis, treatment selection, and follow-up~\citep{bercovich2018medical}.
Compared to the other three most common medical imaging modalities used in the current clinical practice [i.e., Radiography (X-ray), Computerized tomography (CT), and Magnetic resonance imaging (MRI)], Ultrasound (US) imaging has the advantage of being noninvasive, low-cost, portable, and free of ionizing radiation~\citep{izadifar2017mechanical}.
These merits make it particularly suitable for some clinical needs, such as image-guided interventions~\citep{fichtinger2008robotic,neubach2009ultrasound, adebar20143} and obstetric applications~\citep{carneiro2008detection, wright2019complete}. 
\revision{
In October 2021, $0.79$ million US examinations were performed in England, whereas there were $0.52$ million CT scans and $0.31$ million MRI scans~\citep{england2016diagnostic}.}

\par
However, regarding traditional free-hand US examinations, substantial experience and visuo-tactile skills are required for achieving high-quality US images~\citep{jiang2020automatic}. These factors limit the utilization of US in clinical applications requiring reliable biometric measurements or repeatable images for monitoring lesions. To obtain high-quality images, sonographers need to maintain the probe with proper pressure and adjust the probe orientation for optimal acoustic windows. To overcome \secRevision{intra- and inter-operator variations}, the robotic US system (RUSS) has been gaining attention for two decades.
To illustrate the increased interest about RUSS, the number of related \secRevision{peer-reviewed} publications in each year and cumulative years are depicted in Fig.~\ref{Fig_publication_numbers}. For individual years, the number of publications has grown from $1,020$ in \revision{the year $2001$} to $15,500$ in \revision{the year $2022$}. 
The accumulated number of publications exponentially increased to $125,110$ from 2001 to 2022. 

\par
This dramatic rise in interest can be attributed to three distinct communities: engineers, clinicians, and entrepreneurs~\citep{attanasio2021autonomy}. The need from clinicians for high-quality images and efficient and easy-to-use RUSS stimulates the development of RUSS by engineers. Due to the considerable economic benefits, entrepreneurs are motivated to develop prototypes and market them \revision{ \footnote{https://www.adechotech.com/}\textsuperscript{,}\footnote{https://en.mgi-tech.com/}\textsuperscript{,}\footnote{https://www.bkmedical.com/}.}
To assist in combating global pandemics (e.g., COVID-19 and Ebola), the demand for intelligent systems and robotics is boosted extensively in the fields of disease prevention, screening, diagnosis, treatment, home care, etc.~\citep{yang2020combating, khamis2021robotics, di2021medical}.
RUSS has been investigated to remotely or autonomously perform US tests for early detection and diagnosis~\citep{zemmar2020rise,tsumura2021tele}.
Deploying RUSS in hospitals enables the separation of patients and sonographers, hence lowering the risks of virus transmission between patients and medical staff.


\par
This paper is motivated by the desire to assist both robotic US technicians and clinicians. For \final{roboticists}, we provide a comprehensive summary of enabling technologies (i.e., compliant force control and path planning) that are \secRevision{commonly} needed for a variety of applications. In addition to the enabling technologies, \revision{the advanced solutions developed by integrating additional techniques (e.g., surface registration, visual servoing, and image segmentation) are summarized to demonstrate the potential of RUSS for addressing real-world challenges (e.g., tissue motion and deformation).} Using these techniques, clinicians and technicians can further consider how RUSS can assist them in addressing particular clinical needs by sensibly integrating the different techniques together. This will help to bridge the gap between medical and technology research.

\begin{figure*}[ht!]
\centering
\includegraphics[width=0.98\textwidth]{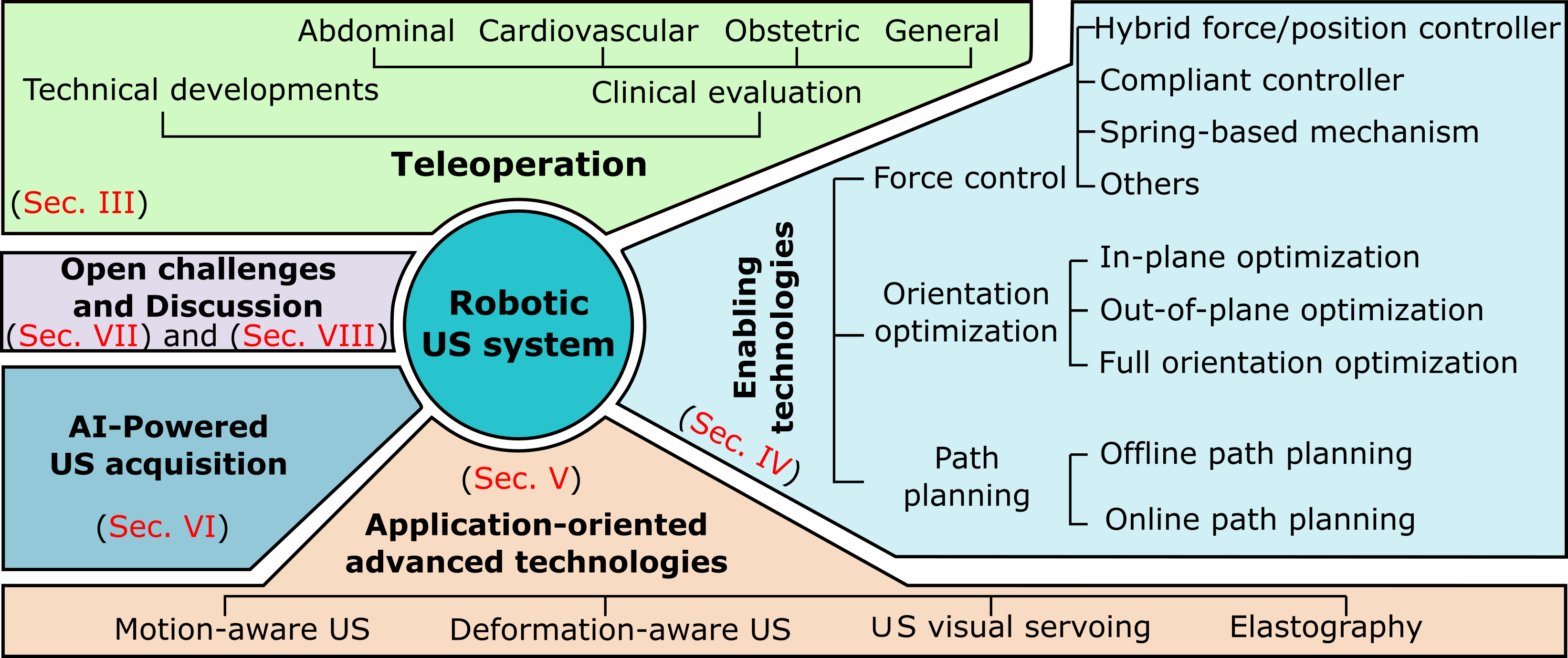}
\caption{Important components of the robotic US and the organizational structure of this work. \secRevision{
Sections~\ref{sec_IV_enable}, \ref{sec_V_Adv}, and \ref{sec_VI_AI} are organized based on the level of technical complexity. By incorporating other advanced techniques (e.g., computer vision, advanced sensing, data fusion, and AI) to fundamental enabling technologies, the level of technical complexity is increased, as is the potential for dynamic and adaptive US imaging in real scenarios.}
}
\label{Fig_overview_structure}
\end{figure*}

\par
Prior to this survey, there were some reviews that summarized the development of RUSS~\citep{priester2013robotic, elek2017robotic, swerdlow2017robotic, salcudean2022robot}. \revision{Recently, Salcudean~\emph{et al.} discussed the roles robotics play in the acquisition of medical images, including US, endoscopy, X-ray, optical coherence tomography, and nuclear medicine~\citep{salcudean2022robot}.}
Specific to RUSS, Von Haxthausen~\emph{et al.} provided a systematic summary of recent publications between $2016$ and $2020$~\citep{von2021medical}. Li~\emph{et al.} focused on the development of autonomous RUSS~\citep{li2021overview}. These two surveys categorize literature based on the level of automation; in contrast, this article emphasizes the connection between the potential clinical applications and enabling techniques. In addition, some novel concepts of application-oriented techniques (e.g., motion-aware~\citep{jiang2021motion} and deformation-aware~\citep{jiang2021deformation} characteristics) have not been discussed before. However, they are important to further pave the way for applying RUSS in real scenarios. Due to the fast development of artificial intelligence (AI), learning-based RUSS is emerging to automatically perform specific US examinations~\citep{burke2020learning, baumgartner2017sononet}.
Li~\emph{et al.} also noted this trend and mentioned the AI-based RUSS as one of the future directions~\citep{li2021overview}. 
Nevertheless, learning-based RUSS solutions have not been systematically discussed yet. 
Therefore, a comprehensive survey article covering these new trends of RUSS will be helpful for \final{roboticists} to quickly and systematically learn the key knowledge of RUSS, as well as for clinicians to comprehend how the robot benefits their specific clinical needs. 
Regarding future development for RUSS, we discussed some open challenges and promising perspectives to inspire the research community and other stakeholders. 


\section{Materials and Methods}
\subsection{Searching Policy}
\par
In order to provide an objective view of the development of robotic US imaging over the last two decades, we \revision{carried out} an extensive search of RUSS on the Web of Science and \revision{google scholar}. The search term was ``(remote OR teleoperat*) AND (ultrasound OR US OR ultrasonography OR echography)", and ``robot* AND (ultrasound OR US OR ultrasonography OR echography) AND (Imaging OR screening OR scan* OR acquisition* OR servoing)". To further narrow the most relevant and most impactful articles, the titles and abstracts were carefully reviewed to exclude the articles that were (a) not focusing on the medical domain, (b) not using robotic imaging adjustment or optimization, or (c) not employing traditional 2D/3D probes. This excludes papers using endocavitary probes~\citep{melodelima2004combination} for cardiology and prostate applications. \revision{Finally, among similar articles, the most representative ones (the newest or most cited) were selected.}

\subsection{\secRevision{Technological Developments in RUSS}}~\label{sec_II_B_method_devlop}

\par
\revision{Skilled sonographers are often in shortage, particularly in rural areas.}
\revision{To allow accurate adjustment of US acquisition parameters} and address the unbalanced distribution of healthcare resources across nations and regions, teleoperated RUSS solutions have been developed over the past two decades (see \textbf{Section~\ref{sec_III_tele}}). For such systems, the operations are fully carried out by experts via teleoperation techniques; thereby, remote experts take the responsibility of robotic acquisition. 
\secRevision{To improve the level of autonomy of RUSS, quite a large number of RUSS solutions have been proposed for different applications in the past decades.} To \secRevision{review} the key characteristics of autonomous RUSS, we first summarize the existing articles in terms of enabling technologies, namely three key acquisitions parameters: contact force (\textbf{Section~\ref{sec_IV_A_enable_force}}), probe orientation (\textbf{Section~\ref{sec_IV_B_enable_orj}}), and scan path (\textbf{Section~\ref{sec_IV_C_enable_path}}). By precisely controlling these parameters, the accuracy and reproducibility of US imaging can be improved~\citep{gilbertson2015force}.

\par
In addition, more advanced techniques need to be developed to tackle additional practical complications occurring in clinical routines, e.g., patient movement and probe pressure-induced deformation.
In this article, we featured four advanced techniques: 1) motion-aware US imaging (\textbf{Section~\ref{sec_V_A_Adv_motion}}), deformation-aware US imaging (\textbf{Section~\ref{sec_V_B_Adv_def}}), US visual servoing (\textbf{Section~\ref{sec_V_C_Adv_servo}}), and elastography imaging (\textbf{Section~\ref{sec_V_D_Adv_Elas}}). 
Sonographers often need to search for standard examination planes for biometric measurement and diagnosis. It is a time-consuming and non-repeatable process, even for experienced sonographers, due to the noisy US images and tissue motion. Benefiting from the development of artificial intelligence, and in particular deep learning, \secRevision{the area of medical image processing has achieved phenomenal success~\citep{maier2022metrics, bass2021ultrasound, hesamian2019deep, shan2021ultrasound, ronneberger2015u}.} Learning-based image processing techniques lead to accurate and robust understandings of US images, which further enables training RUSS to learn both manipulation skills and clinical knowledge directly from human sonographers. We summarize the most \secRevision{recent developments in learning-powered RUSS} (\textbf{Section~\ref{sec_VI_AI}}), aiming to automatically search for specific anatomy or navigate a probe to visualize standard US planes. Finally, we discuss the open challenges and provide a few potential directions for future developments \textbf{Section~\ref{sec_VII_furtu}}. The important components of robotic US and the organization structure of this article are depicted in Fig.~\ref{Fig_overview_structure}. 
\secRevision{By incorporating additional techniques to fundamental enabling technologies, the level of technical complexity is increased from Section~\ref{sec_IV_enable} to Section~\ref{sec_VI_AI}. In this way, we would like to highlight our strategy to inspire the community to achieve the ultimate goal of developing an intelligent robotic sonographer that can collaborate with human sonographers to improve diagnostic and intraoperative imaging in real scenarios.
}

\newpage
\section{Teleoperation in RUSS}~\label{sec_III_tele}
\par
Teleoperation allows operators to remotely carry out certain tasks. Due to the development of networks, multimedia, and communication technologies in the past decades, teleoperation has become one of the most mature techniques for reforming modern medical procedures~\citep{ye2021feasibility}. The main characteristic of teleoperation is that the robot's motion is controlled by operators. This is important for obtaining regulatory approval. The most successful representative is da Vinci from Intuitive Surgical, which has become the clinical standard of minimally invasive surgery for a wide range of surgical procedures~\citep{esposito2021robotics}. 
\secRevision{Regarding teleoperated RUSS}, it has been seen as a solution for work-related musculoskeletal disorders of sonographers~\citep{brown2003work, fang2017force}. In addition, separating operators from patients reduces the risk of transmitting pandemics (e.g., Covid-19)~\citep{tsumura2021tele, ye2021feasibility}.
This section summarizes the technical and clinical contributions of remote RUSS, respectively.  

\begin{figure*}[ht!]
\centering
\includegraphics[width=0.93\textwidth]{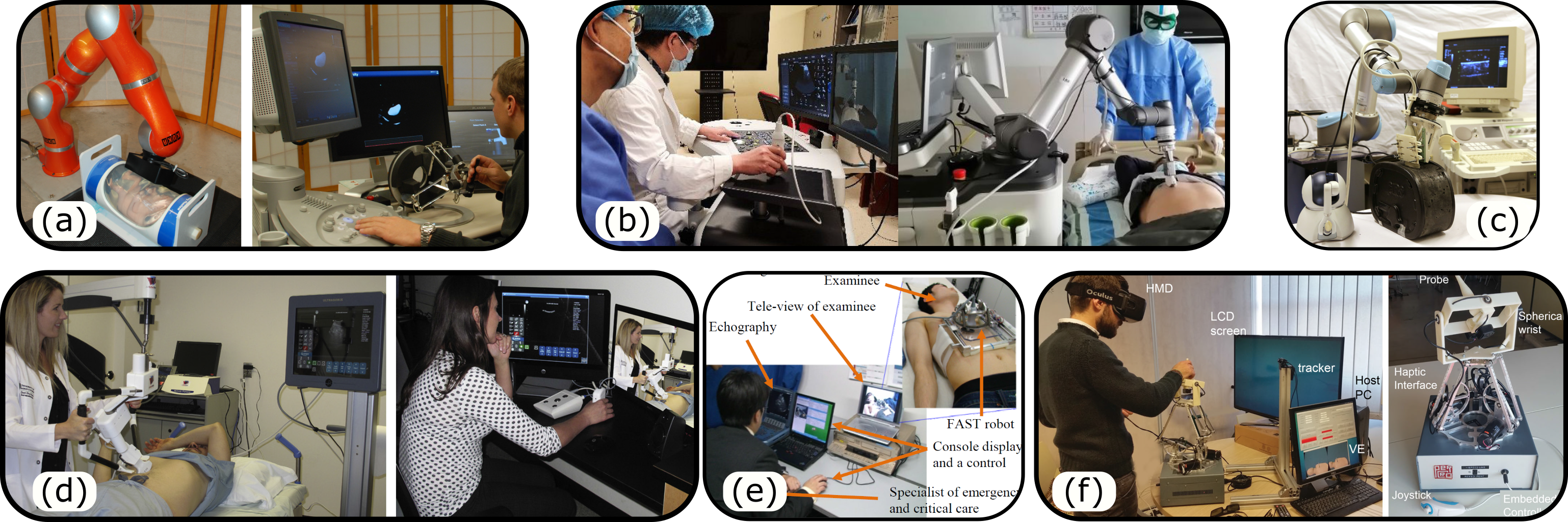}
\caption{Representative teleoperated RUSS. (a) a commercial 6 DOF haptic device (Omega $6$) used to control an industrial robotic arm~\citep{conti2014interface}. (b) 5G-Based robot-assisted remote US system for COVID-19 diagnosis~\citep{ye2021feasibility}. (c) a RUSS developed based on an Omni haptic device and a UR5 robotic manipulator~\citep{mathiassen2016ultrasound}. (d) MELODY System used for remote abdominal examinations~\citep{adams2017initial}. (e) a custom-designed passive mechanism emphasizing portability and attachability for emergency cases~\citep{ito2010portable}. (f) a teleoperated RUSS incorporating VR for remotely controlling the probe in an immersive environment~\citep{filippeschi2019evaluation}.
}
\label{Fig_tele_multiple_systems}
\end{figure*}

\subsection{Technical Developments}~\label{sec_III_A_tele_techni}
\par
\secRevision{Teleoperated RUSS often consists} of three individual components: 1) an expert console, 2) a patient-side manipulator (PSM) used to maneuver a US probe, and 3) a software control system mapping the movement made by experts to the PSM. The teleoperated RUSS allows sonographers to manually, \final{unconstrainedly}, and safely control the probe motion onto the patient via the PSM. 
\secRevision{
Teleoperated systems are also utilized on-site because robotic systems can overcome human limits in manipulation and perception by adding dexterity and precision. A common example is da Vinci, which is often employed on-site~\citep{haidegger2011surgery}. 
}

\subsubsection{Robotic Mechanism}
\par
\revision{
In $1999$, Salcudean~\emph{et al.} designed a \secRevision{six degree of freedom (DOF)} lightweight mechanism with limited force capability for teleoperated RUSS~\citep{salcudean1999robot}. Due to the need for a large orientation workspace, a parallelogram linkage was employed to decouple the orientation and translation in their final design, achieving the control resolution of $0.1~mm$ for translation and $0.09^{\circ}$ for rotation. Similarly, Lessard~\emph{et al.} designed the PSM in parallel structure in order to have enough workspace~\citep{lessard2007new}.}
\revision{
Masuda~\emph{et al.} designed a 6-DOF mechanism consisting of gimbals, pantograph and slide mechanisms, which weighed $3.3~kg$~\citep{masuda2001three}.}
To guarantee the safety of patients, there are four sensors symmetrically \secRevision{deployed} around the probe to monitor real-time force.

\par
In addition, a number of soft mechanisms were developed for force-sensitive applications, e.g., obstetric examinations, to strictly limit the maximum US probe pressure. Vilchis~\emph{et al.} proposed a \revision{cable-driven nonrigid remote robot~\citep{vilchis2003new}}. This system has been used on $100$ patients with abdominal aortic aneurysm (AAA) at a distance of $1125~km$. Tsumura~\emph{et al.} designed a \revision{passive mechanism using springs for fetal examinations, which \secRevision{can} prevent excessive contact force~\citep{tsumura2020robotic}.} Besides, a portable and attachable robotic system has been designed by Ito \emph{et al.}~\citep{ito2010portable} [see Fig.~\ref{Fig_tele_multiple_systems} (e)]. In the same direction, Vieyres~\emph{et al.} proposed a 4-DOF light mechanism with \secRevision{3-DOF} rotation and \secRevision{1-DOF translation} in probe centerline~\citep{vieyres2003teresa}. Then, they updated \secRevision{the design of the portable RUSS to allow all 6-DOF motions} using serial mechanism~\citep{vieyres2006tele}. The portable RUSS \secRevision{is} easily used by paramedics, which makes it ideal for use in emergency medical circumstances. Nevertheless, owing to the need of the compact structure, portable RUSS typically have restricted working space.
\revision{Since mechanical design is beyond the scope of this survey's primary focus on imaging acquisition, we refer readers to two comprehensive review articles with mechanical designs for RUSS~\citep{priester2013robotic, avgousti2016medical}. 
}

\par
To reduce the cost of RUSS, commercial robotic manipulators e.g., Universal Robot \secRevision{(University robot, Denmark)} and Franka Emika Panda \secRevision{(Franka Emika GmbH, Germany)} are often used as PSM~\citep{filippeschi2021kinematic, mathiassen2016ultrasound}  [see Fig.~\ref{Fig_tele_multiple_systems} (b) and (c)]. 
\revision{It is noteworthy that another typical standard robotic arm KUKA LBR iiwa (KUKA Robotics GmbH, Germany), with integrated joint torque sensors, is also commonly employed as a PSM~\citep{schreiter2022towards, schreiter2022multimodal}.} 
HIPPOCRATE is a representative of teleoperated RUSS \secRevision{developed using} a serial industrial robotic arm~\citep{degoulange1998hippocrate,pierrot1999hippocrate}.

\subsubsection{\secRevision{Shared Autonomy in Teleoperated RUSS}}
\par
To fully take advantage of the stability and accuracy of robotic techniques, Abolmaesumi~\emph{et al.} proposed a shared autonomy strategy between an expert and an image servo~\citep{abolmaesumi2001user}. The in-plane three DOFs were controlled by visual servoing to automatically center the carotid artery in cross-sectional images, while the other three DOFs were teleoperated by an expert. In this case, the image servo can provide pixel-by-pixel control accuracy and further mitigate the negative influence of human tremor. To keep the tissue of interest always visible in the image and give more flexibility to the expert, Li~\emph{et al.} and Krupa~\emph{et al.} shared all four (in-plane and out-of-plane) DOFs of a lightweight body-mounted mechanism between the visual servoing algorithm and a human operator via teleoperation~\citep{li2012maintaining, krupa2014robotized}. The visual servoing technique has also been widely used in autonomous RUSS to estimate and compensate for the motion of internal organs~\citep{krupa2009real}, visualize and track the object of interest~\citep{mebarki20102,krupa2007real}, and improve the image quality by optimizing the acoustic windows~\citep{chatelain2015optimization}, etc. Please refer to Section~\ref{sec_V_C_Adv_servo} for more details.


\subsubsection{User Interface}
\par
\revision{
Masuda~\emph{et al.} employed two joysticks to remotely control the three-dimensional rotation and translation individually of the PSM~\citep{masuda2001three}.}
Yet, this manner differs from how experts conduct conventional US examinations. To \secRevision{enhance} the intuitiveness of the interaction, a dummy probe is frequently utilized to intuitively control PSM from the expert console~\citep{arbeille2014teles}. A gyroscope was installed within the dummy probe so that it could track the motion of the expert~\citep{georgescu2016remote}. To improve the accuracy of the motion estimation, some mature techniques, such as optical and electromagnetic tracking can be utilized.
As the use of a dummy probe allows experts to conduct US examinations as usual, RUSS can reduce training time and increase examination efficiency.
However, the lack of force feedback on expert side may hinder the clinical acceptance.
To tackle this problem, Martinelli~\emph{et al.} employed a haptic control system that rendered contact force in three dimensions~\citep{martinelli2007robot}.
Conti~\emph{et al.} employed a commercial 6-DOF haptic device (Omega $6$) to reflect the contact force in six dimensions~\citep{conti2014interface} [see Fig.~\ref{Fig_tele_multiple_systems} (a)]. 
\revision{Recently, Naceri~\emph{et al.} directly deployed two 7-DOF Franka Emika Panda~\citep{naceri2022tactile}, one of which was used at expert console with force feedback, and the other one used at patient side to precisely reproduce the movements of the experts.}

\par
Benefiting from the development of virtual reality (VR) techniques, a VR simulator was designed as a new type of interface for teleoperated \secRevision{RUSS}~\citep{filippeschi2019evaluation} [see Fig.~\ref{Fig_tele_multiple_systems} (f)]. Compared to traditional joysticks or other haptic devices, an immersive experience can be achieved using VR simulators, which could intuitively visualize the remote scenes in 3D. The initial evaluation of \secRevision{a VR simulator} has been performed by $12$ experienced sonographers and the results suggest that the immersive simulator could be used for teleoperated RUSS~\citep{filippeschi2019evaluation}. 
\secRevision{A deeper discussion about human-robotic interaction studies will be beyond the focus of this paper. To inspire further research incorporating novel human-machine interfaces to improve the efficiency, intuitiveness, and robustness of teleoperated RUSS, we refer readers to two comprehensive surveys on interface approaches~\citep{kumar2022human, mahmud2020interface}. Specific to medical applications, Abdelaal~\emph{et al.} provided a crucial review of interfaces that have been used or tested in vivo~\citep{abdelaal2020robotics}.}

\subsection{Clinical Feasibility Evaluation}~\label{sec_III_B_tele_clinical}
\par
Teleoperated RUSS can fully utilize the advanced knowledge of experts. Compared to autonomous RUSS, teleoperated RUSS is easier to be certified for clinical use due to the fact that all diagnostic decisions and scan trajectory are made by experts. To achieve this objective, clinical studies have been performed using different teleoperated RUSS for a number of examinations. Clinical evaluations of existing teleoperated RUSS solutions have been categorized according to their clinical applications as TABLE-\ref{Table_clinical_evaluation_tele}.

\begin{table*}[!ht]
\normalsize 
\centering
\caption{Representative Clinical Feasibility Evaluations of Teleoperated RUSS}
\label{Table_clinical_evaluation_tele}
\resizebox{\textwidth}{!}{
\begin{tabular}{cccccc}
\noalign{\hrule height 1.2pt}
Systems & Medical applications & Published year & No. of participants & Distances & Country \\ 
\noalign{\hrule height 1.0 pt}
\cite{arbeille2003echographic}  & General abdominal organs   &2003   &20   &20--50~km  &France\\
\cite{arbeille2007use}          & General abdominal organs   &2005   &81   &20~km  &France\\
\cite{adams2017initial}          & General abdominal organs   &2016   &18   &2.75~km  &Canada\\
\cite{martinelli2007robot}          & Abdominal aorta   &2007   &58   &1000~km  &France\\
\cite{arbeille2014teles}          & Cardiovascular   &2014   &41   &10~km  &France\\
\cite{boman2009remote}          & Cardiovascular   &2009   &17\&31   &80\&135~km  &Sweden\\
\cite{boman2014robot}          & Cardiovascular   &2014   &38   &217~km  &Sweden\\
\cite{sekar2007telecardiology}          & Pediatric cardiology   &2007   &102   &193~km  &India\\
\cite{sengupta2014feasibility}          & Carotid   &2014   &1   &trans-Atlantic  &America\\
\cite{avgousti2016cardiac}          & Cardiovascular   &2016   &2   &unknown  &Cyprus\\
\cite{arbeille2005fetal}          & Obstetric imaging   &2005   &29   &1700~km  &France\\
\cite{adams2018crossover}          & Obstetric imaging   &2018   &30   &adjacent room  &Canada\\
\cite{georgescu2016remote}          & General cases   &2016   &300   &50~km  &France\\
\cite{jiang2023application}   & \secRevision{General cases}     &\secRevision{2023}   &\secRevision{22} &\secRevision{adjacent room} &\secRevision{China}\\
\noalign{\hrule height 1.2 pt}
\end{tabular}
}
\end{table*}


\subsubsection{Abdominal Imaging}
\par
The abdomen is often examined using US images, which is one of the primary focuses of teleoperated RUSS. To validate the feasibility and diagnostic accuracy of such systems, Arbeille~\emph{et al.} evaluated a preliminary version of a teleoperated RUSS for general abdominal imaging on $20$ patients~\citep{arbeille2003echographic}. The expert was in a room at some distance ($20-50~km$) from the patient's site. The time delay between experts and the PSM 
was less than $0.1~s$ using ISDN (terrestrial) telephone lines and less than $0.5~s$ using satellite links. To evaluate the performance, the authors validated their approach on four different groups of organs. The results demonstrated that the expert could image the main views (longitudinal and transverse) of the liver, gallbladder, kidneys, aorta, pancreas, bladder, and uterus on the patient. Only the heart and spleen were not identified in two and four of the $20$ cases, respectively. The experiments also showed that sonographers can master the teleoperated RUSS in less than $3$ hours, while the examination time ($27\pm7~min$ for three or four organs) was approximately $50\%$ longer than the traditional US examination. 

\par
In a following study, Arbeille~\emph{et al.} further compared the performance of robotized and conventional US examinations on $87$ patients examined in the emergency department at the Tours University in France~\citep{arbeille2007use}. The results demonstrated that each organ (e.g., liver, gallbladder, pancreas, kidney) can be correctly imaged by a robotized system in between $91\mbox{-}100\%$ of cases compared with the conventional US examinations. In addition, the mean visualization score for the teleoperated RUSS was $87.4\%$ for the abdomen, while there were no false diagnoses made in this study~\citep{arbeille2007use}. In another clinical evaluation, Adams~\emph{et al.} also assessed the feasibility of performing adult abdominal US examination using a remote RUSS on $18$ patients in the University of Saskatchewan, Canada~\citep{adams2017initial}. Telerobotic examinations were successful in $92\%$ of the examinations on various abdominal organs (given the organs were sufficiently visualized on the conventional examination); 
five pathological findings were identified on both modalities, three and two findings were only identified using conventional and telerobotic system, respectively. Furthermore, they reported that all participating patients were willing ($89\%$ were strongly willing and the remaining $11\%$ were willing) to have another telerobotic examination~\citep{adams2017initial}.

\par
Martinelli~\emph{et al.} carried out a study on $58$ patients with a focus on the aorta~\citep{martinelli2007robot}. The examination results demonstrated that all aneurysm cases were correctly detected by both conventional scans and the teleoperated RUSS. Furthermore, the quantitative results show that the diameter of the patient's aorta can be accurately measured. The interobserver correlation coefficient was $0.98$ and the difference in measurement was less than $4~mm$ in $96.3\%$ cases. In addition, the examination duration (mean$\pm$SD) of the teleoperated system and traditional examinations are $17\pm8~min$ and $12\pm7~min$, respectively. Finally, they also reported that the acceptability of patients was $84\pm18\%$, which is similar to the result in~\citep{adams2017initial}.

\subsubsection{Cardiovascular Imaging}
\par
Compared with general abdominal organs, cardiac examinations are considered more technically demanding procedures. Regarding echocardiography, the clinical needs include the visualization and evaluation of the four cardiac chambers, measurements of aortic flow, and the identification of mitral, tricuspid, or aortic valve leaks or aortic stenosis~\citep{arbeille2014teles}. To successfully perform tele-echocardiography, the probe was held by a \secRevision{3-DOF} robotic arm providing three orthogonal rotations, and then, the robotic arm was fixed to a motorized plate for obtaining translational movements~\citep{arbeille2014teles}. The results on $41$ cardiac patients demonstrated that similar measurements can be achieved in most cases ($93\%\mbox{-}100\%$). 
Among the $71$ valve leaks or aortic stenosis patients, $61$ ($86\%$) were successfully detected using tele-echocardiography and there was no false-positive diagnosis reported.

\par
Boman~\emph{et al.} also carried out a similar study on cardiovascular examination in Sweden~\citep{boman2009remote}. The evaluations were carried out in three different stages. In stage 1, there were $27$ patients in a different place than sonographers with a distance of $80~km$. Regarding the other two stages, a total of $31$ subjects were recruited in a place at $135~km$ from the experts. The results indicate that real-time echocardiographic examinations are possible~\citep{boman2009remote}. Boman~\emph{et al.} compared the tele-echocardiography examination with the standard of care referral approach in terms of time and diagnosis~\citep{boman2014robot}. $19$ patients were randomized to remote consultation and imaging, and $19$ to the standard of care consultation. The results demonstrated that the processing time was significantly reduced in the remote one (only $26.5$ days vs $114$ days for the standard one). Therefore, compared with the standard of care approach, patients were more satisfied with the remote consultation strategy, which offered an increased rapidity of diagnosis and the likelihood of receiving faster patient management~\citep{boman2014robot}.

\par
In 2007, Sekar~\emph{et al.} evaluated tele-echocardiography examination in the diagnosis of congenital heart diseases in pediatric populations~\citep{sekar2007telecardiology}. In this 3-year study, $102$ pediatric telecardiology examinations were performed between a tertiary care cardiac center and a remote rural hospital located $193~km$ away. Pathology was ruled out in $50$ children by tele-echocardiography. In addition, heart lesions were identified in $52$ children and $30$ among them required surgery. By using teleoperation techniques, the total cost for such remote care can be controlled under $90$ USD, which becomes considerable for most developing areas~\citep{sekar2007telecardiology}. Sengupta~\emph{et al.} further validate the feasibility of long-distance (trans-Atlantic) telerobotic US scans for vascular examinations~\citep{sengupta2014feasibility}. The results showed that the procedure to localize the remote probe along the short axis of the carotid artery took less than $60~s$ and an examination could successfully be conducted in $4~min$. Avgousti~\emph{et al.} employed 4G wireless networks in order to reduce the time delay for live tele-echography~\citep{avgousti2016cardiac}. \secRevision{However, it is also important to note that the communication stability and potential signal interference may lead to uncertainty.}

\subsubsection{Obstetric Imaging}
\par
Obstetric imaging is also one of the most frequent applications of US examination in clinical practice. From the beginning phase to the birth of infants, more than five fetal examinations are carried out and such examinations are important to evaluate the health of both \revision{fetuses} and pregnant women~\citep{arbeille2005fetal}. To assess the feasibility of teleoperating fetal US examinations in pregnant women, Arbeille~\emph{et al.} carried out a study on $29$ pregnant women in an isolated hospital $1700~km$ away using both conventional and teleoperation examinations~\citep{arbeille2005fetal}. The results demonstrated that the biometric parameters, placental location, and amniotic fluid volume can be correctly measured in most cases ($93.1\%$) using a teleoperated RUSS. Only in two cases, femur length could not be correctly measured. The mean duration of US examination of the remote \secRevision{examinations} ($18~min$) was longer than that of conventional examinations ($14~min$). 

\par
Another study with a similar objective was presented by Adams~\emph{et al.} on $30$ patients in Canada~\citep{adams2018crossover}. In this study, the results indicated that there was no statistically significant difference between teleoperated RUSS and conventional measurements of overhead circumference, biparietal diameter, or single deepest vertical pocket of amniotic fluid; however, there were slight differences in the measures of abdominal circumference and femur length. Besides, $80\%$ of the fetal structures could be sufficiently acquired by the telerobotic system (range, $57\%–100\%$ for each patient). Finally, a survey of participants shows that $92\%$ patients are willing to have another telerobotic examination in the future. The aforementioned studies demonstrated the feasibility of using teleoperation to remotely carry out fetal US examinations while keeping comparable biometric measurements as precise as the conventional approach. 

\subsubsection{General \secRevision{Applications}}
Georgescu~\emph{et al.} reported the usability of a teleoperation system for general applications over one year~\citep{georgescu2016remote}. In total $300$ patients were involved: $138$ supra-aortic vessels, $68$ abdomen, $33$ thyroid, $30$ lower limb vein, $20$ pelvis, $7$ kidneys, $3$ small parts, and $1$ obstetrics. The reported average duration of a teleoperation examination was $24\pm5~min$ over all $300$ examinations. In addition, the results showed that the use of teleoperation in the general medicine practice significantly reduced the waiting time (save several days) for patients, and similar information as conventional US examinations was achieved. It also contributed to saving costs for the healthcare system and facilitating earlier treatment of conditions, potentially leading to improved patient outcomes and less time in care facilities~\citep{georgescu2016remote}. \secRevision{Most recently, a teleoperated RUSS was tested on $22$ Covid-19 patients, and they concluded that teleoperated RUSS can be used to diagnose common abdominal, vascular, and superficial organ pathologies with acceptable accuracy~\citep{jiang2023application}.}











\section{Enabling Technologies for Autonomous RUSS}~\label{sec_IV_enable}
\par
\secRevision{Recently}, interest in autonomous RUSS has increased relatively to teleoperated RUSS. Autonomous RUSS has the potential to achieve standardized and reproducible \revision{US acquisitions}. RUSS solutions further release sonographers from burdensome manipulation tasks and allow them to focus on diagnosis, requiring deep anatomical and physiological knowledge. 

\par
The move of the research community toward autonomous RUSS has also proposed novel scientific questions, \revision{which defined important and exciting challenges. To develop autonomous RUSS, we first need to understand how human sonographers perform US scans. In this paper, we call this process the recovery of the ``language of sonography". \revision{The community has not investigated this consciously, but this path can be traced throughout the analysis of the state of the art.}
The adjustment of contact force, probe position and orientation for optimal image acquisition has often been the first focus. Then, \revision{it is also crucial to plan} an appropriate path for covering the area of interest and to compensate for the potential motion and deformation of the target anatomy during imaging.}
These points will be discussed \revision{explicitly} in the following sections in more detail when we review some of the most relevant states of the art.


\par
In this section, three fundamental techniques used in RUSS are elaborated: 1) compliant control used to apply and maintain a given contact force between US probe and patients, 2) orientation optimization to determine the appropriate probe orientation for a given scan (often orthogonal to the contacted surface) and 3) path planning to best localize and visualize the anatomy of interest.

\subsection{Force Control Approaches}~\label{sec_IV_A_enable_force}
\par
Due to the inherited characteristic of US imaging, a certain contact force between a US probe and human tissues is required to optimize acoustic coupling, thereby achieving high-quality US images. It is challenging for human operators to maintain a constant force during US scans. The varying force will result in non-homogeneously deformed US images. Thus, a dedicated force controller is needed to maintain the contact force during scans. Furthermore, such a controller is also crucial for guaranteeing the safety of patients \revision{by preventing excessive force.} 
\revision{
Depending on the target tissues, the acceptable contact force is less than approximately $20~N$~\citep{tsumura2021tele}. In the meanwhile, a small force (less than $1.2~N$) is commonly considered as not being in complete contact with the skin~\citep{dhyani2017pilot}.}
\revision{It is noteworthy that this subsection only summarized the force control approaches (both software and hardware-wise) that have been used for developing RUSS. A more general and comprehensive summary of force control can refer to~\citep{haidegger2009force, zeng1997overview}.}

\begin{table*}[!ht]
\normalsize 
\centering
\caption{\revision{Representative Force Control Approaches used in RUSS}}
\label{Table_fore_control}
\resizebox{\textwidth}{!}{
\revision{
\begin{tabular}{cccccc}
\noalign{\hrule height 1.2pt}
Method & Reference & Sensor & Robot & Applications & Data\\ 
\noalign{\hrule height 1.0 pt}
\multicolumn{1}{c}{\multirow{5}{*}{\makecell[c]{Hybrid \\Force/Position \\Controller}}}  & \cite{gilbertson2015force}   &1-DOF load cell   & handheld mechanism  &general  &phantom\\
\multicolumn{1}{c}{}   & \makecell[c]{\cite{zhu2000motion};\\\cite{abolmaesumi2002image}}   &6-DOF F/T sensor  &6-DOF customized robot  &carotid  &phantom \& volunteer\\
\multicolumn{1}{c}{}   & \cite{pierrot1999hippocrate}   &6-DOF F/T sensor   &7-DOF PA-10 robot   &cardiovascular  &patient\\
\multicolumn{1}{c}{}   & \cite{ma2021novel}   &built-in torque sensor   &6-DOF UR5e   & vascular structure &phantom\\
\noalign{\hrule height 0.5 pt}
\multicolumn{1}{c}{\multirow{7}{*}{\makecell[c]{Compliant \\Controller}}} & \makecell[c]{\cite{jiang2021autonomous_TIE_vessel, jiang2020automatic_tie_normal, jiang2021deformation}; \\\cite{welleweerd2021out}; \\\cite{hennersperger2016towards}}   &built-in torque sensor   & 7-DOF KUKA iiwa  &\makecell[c]{general \& vessel\\ \&breast}  &phantom \& volunteer\\
\multicolumn{1}{c}{}   & \cite{suligoj2021robust}   &built-in torque sensor   &7-DOF Franka Pamda   & vessel &phantom\\
\multicolumn{1}{c}{}   & \cite{dyck2022impedance}   &torque sensor   &7-DOF DLR MIRO   &general  &phantom\\
\multicolumn{1}{c}{}   & \cite{ning2021autonomic, ning2021force}   &6-DOF F/T sensor   &6-DOF UR3   & spine &phantom \& human\\
\multicolumn{1}{c}{}   & \cite{fang2017force}  &6-DOF F/T \& 1-DOF sensor   &6-DOF UR5   & general &phantom\\
\multicolumn{1}{c}{}   & \cite{wang2022full}   &6-DOF F/T sensor   &6-DOF UR5   &breast  &phantom\\
\noalign{\hrule height 0.5 pt}
\multicolumn{1}{c}{\multirow{6}{*}{\makecell[c]{Spring-based\\mechanism}}}  & \cite{tsumura2020robotic}   &spring   & \makecell[c]{3-DOF linear stage \\\& customized end effector}  &fetal  &phantom\\
\multicolumn{1}{c}{}   & \cite{wang2019analysis} &spring    &8-DOF customized mechanism   & fetal &phantom\\
\multicolumn{1}{c}{}   & \cite{bao2022sapm} &spring    &\makecell[c]{robotic arm\\\&3-DOF customized end-effector}   & fetal &phantom\\
\multicolumn{1}{c}{}   & \cite{housden2021towards}  &spring   &\makecell[c]{17-DOF customized \\mechanism with two arms}    & fetal &phantom \& volunteer\\
\multicolumn{1}{c}{}   & \cite{bao2021constant}  & \makecell[c]{spring \\\& 6-DOF F/T sensor}  &\makecell[c]{robotic arm \\\& customized end effector}  & fetal & phantom\\
\noalign{\hrule height 0.5 pt}
\multicolumn{1}{c}{\multirow{3}{*}{\makecell[c]{Others}}}  & \cite{huang2018fully}   &two 1-DOF load cells & 3-DOF linear stage  &general  &phantom\\
\multicolumn{1}{c}{} & \cite{huang2018robotic}  &two 1-DOF load cells  &6-DOF robotic arm   & general & phantom\\
\multicolumn{1}{c}{} & \cite{ma2022see}   &four laser dist. sensors  &\makecell[c]{7-DOF Franka Pamda\\\& customized end effector}    & general \& lung &phantom\\
\noalign{\hrule height 1.2 pt}
\end{tabular}
}
}
\end{table*}

\subsubsection{Hybrid Force/Position Controller}
\par
The \textbf{traditional hybrid force/position control approaches} are implemented in two decoupled subspaces taking position law and force control law, respectively, into account~\citep{raibert1981hybrid}. Both force and position differences between current values and desired values are fed into the robotic dynamic model to update the manipulator's motion. To apply a constant contact force between a probe and subjects, Gilbertson~\emph{et al.} implemented a hybrid position/force controller for a 1-DOF hand-held RUSS~\citep{gilbertson2015force}. In this study, they simplified the contact model as two interfaces (human-machine and probe-patient) using a set of masses, springs, and dampers. Thereby, the contact force can be dynamically connected to the probe position and velocity by selecting proper interface parameters. 
A similar hybrid position/force method based on an external 6-DOF force/torque (F/T) sensor was designed for 6-DOF RUSS~\citep{zhu2000motion, abolmaesumi2002image}. Their approaches can automatically switch between velocity and force control modes according to the contact condition (free or contact space).


\par
The \textbf{External hybrid force/position control} is also often used in RUSS. The external controller first updates the position based on the force; then, \revision{the positional error is controlled using an internal servo.}
Pierrot~\emph{et al.} used a PI controller to maintain the contact force and a PID controller to continually run the joint position servo loop for a 7-DOF robotic US system~\citep{pierrot1999hippocrate}. \revision{Similarly}, Ma~\emph{et al.} used a PID controller to actively compute the variation of Cartesian position based on the force error; and then used a position controller (provided by the manufacturer) in the inner loop~\citep{ma2021novel}.To limit the negative effect caused by potential force measurement errors, a low-pass filter, and a moving filter were used to smooth the measured force. \revision{The authors claimed that the implementation of such an external force controller is simpler and can be adapted for any kind of robot~\citep{pierrot1999hippocrate}. }

\subsubsection{Compliant Controller}
\par
Regarding the hybrid force/position controller, a position controller is employed either in a sub-space for the traditional ones or in the low-level servoing loop for the external ones. Since the environment is unknown in real scenarios, the position control may result in excessive force to move to the computed positions. To ensure the safety of patients, two compliant control methods (\textbf{impedance controller} and \textbf{admittance controller}) are often used. The dynamic model of compliant controller is described as Eq.~(\ref{eq_impedance_law})~\citep{jiang2021autonomous_TIE_vessel}. 

\begin{equation}\label{eq_impedance_law}
\textbf{F} + \textbf{F}_{ext} = \textbf{K}_m e + \textbf{D} \dot{e} + \textbf{M} \ddot{e}
\end{equation}
where $\textbf{F}$ is the applied force/torque in Cartesian space, $e = (x_d - x_c)$ is the Cartesian position and orientation error between the current pose $x_c$ and the target pose $x_d$, $\textbf{F}_{ext}$ is the desired force/torque, $\textbf{K}_m$, $\textbf{D}$ and $\textbf{M}$ are the stiffness, damping and inertia matrices, respectively.


\par
Based on Eq.~(\ref{eq_impedance_law}), the compliant performance can be achieved in all directions by giving different $\textbf{K}_m$ and $\textbf{D}$, which \revision{enables} safe/soft interactions between RUSS and patients. Regarding Eq.~(\ref{eq_impedance_law}), there are two different interpretations, which are referring to \textbf{impedance control} and \textbf{admittance control}, respectively. For the former one, the pose error is seen as feedback and the computed force and torque are applied to achieve the expected force $\textbf{F}_{ext}$. On the other hand, for an admittance controller, the force applied at the end-effector $\textbf{F}$ is measured as input, while the output is the Cartesian movement. Since admittance control only requires the measurement of external force/torque, it is often used for \revision{low-cost robots} without accurate joint torque sensors, e.g., universal Robots~\citep{fang2017force, ning2021autonomic, ning2021force}. On the contrary, impedance control is more often used when robotic manipulators are equipped with accurate joint torque sensors, e.g., KUKA LBR iiwa~\citep{jiang2021autonomous_TIE_vessel, hennersperger2016towards, welleweerd2021out, jiang2020automatic_tie_normal, jiang2021deformation, virga2016automatic} and Franka Emika Panda~\citep{suligoj2021robust}. 
\revision{When the stiffness of the environment diminishes, the performance of impedance control will decrease due to friction and unmodeled dynamics, while the performance of admittance control will increase~\citep{ott2010unified}. Therefore,} admittance control \revision{could} achieve better performance on soft tissues, while impedance control \revision{could} be more suitable for stiff tissues.

\begin{figure*}[ht!]
\centering
\includegraphics[width=0.90\textwidth]{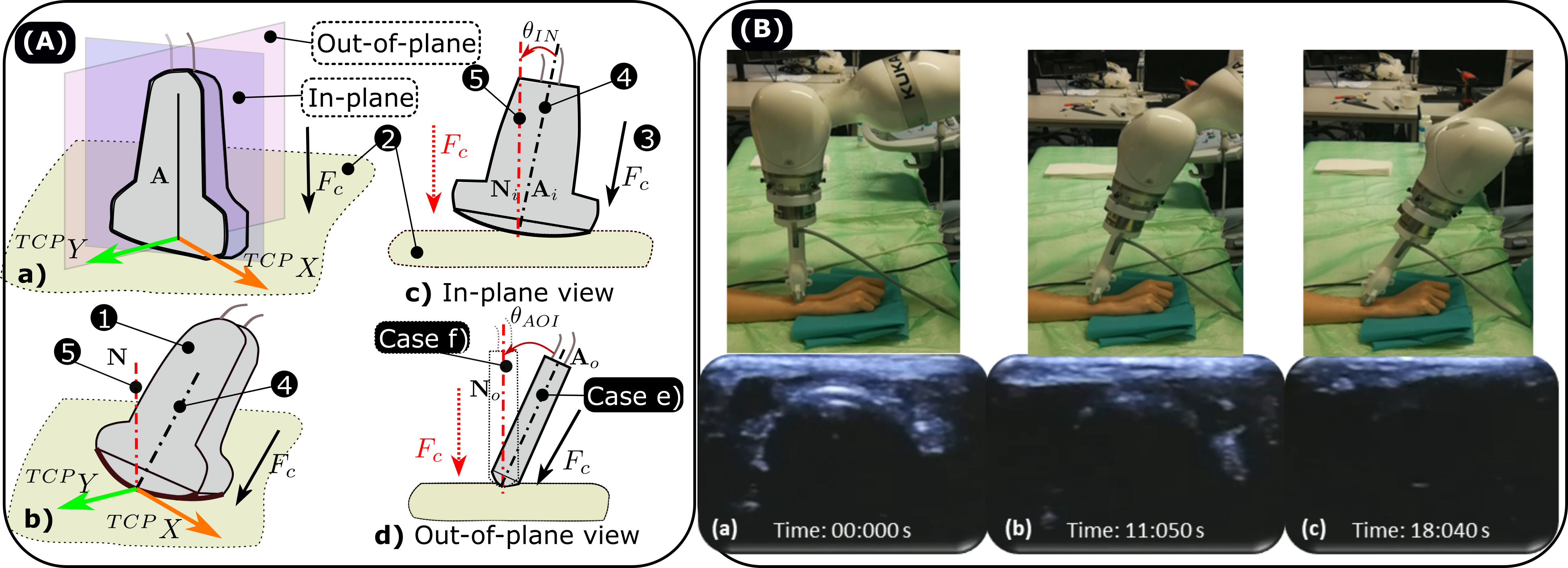}
\caption{Impact of the probe orientation on US images. \textbf{A}~\citep{jiang2020automatic}: 1: US probe; 2: Surface; 3: Contact force $F_c$; 4: probes central axis $\mathbf{A}$; 5: Normal direction $\mathbf{N}$. TCP refers to the tool center point. a) and b) describe ideal and non-ideal probe orientation in 3D, respectively; c) and d) are the in-plane and out-of-plane view, respectively; \revision{$\theta_{AOI}$ and $\theta_{IN}$ are the angle offsets between $\mathbf{A}$ and $\mathbf{N}$ in the out-of-plane and in-plane views, respectively. \textbf{B}~\citep{jiang2020automatic_tie_normal}: intuitive illustration of resulting US images of volunteer's radius bone when the probe tilted around $0$, $15^{\circ}$ and $25^{\circ}$, respectively.}
}
\label{Fig_searchingpolicy}
\end{figure*}

\subsubsection{Spring-based Mechanism}
\par
Since some clinical applications, e.g., fetal examination, are really sensitive to the applied force during US examinations, Tsumura~\emph{et al.} proposed a spring-based mechanism to maintain the contact force and passively adjust the probe pose with respect to the constrained surface~\citep{tsumura2020robotic}. Compared to the aforementioned sensor-based controllers, the \textbf{passive mechanism} can apply a constant force quickly and safely, especially in unstructured environments. Wang~\emph{et al.} proposed a spring-loaded ball clutch to limit the maximum contact force~\citep{wang2019analysis}. In normal cases, the detent structure is in its engaged position with ball restricted by a preloaded compressed spring. Once excessive force occurs, the ball comes out from the detent hole. Thus, the involved clutch joint will lose the function of transmitting torque~\citep{wang2019analysis}. In these ways, the maximum contact force of such mechanisms can be mechanically limited to $10~N$~\citep{tsumura2020robotic} and $21.98\pm0.96~N$~\citep{wang2019analysis}. Yet, this approach cannot precisely and dynamically control the contact force.

\par
To address this challenge, Housden~\emph{et al.} extended their work~\citep{wang2019analysis} by integrating a customized multi-axis F/T sensor to allow \textbf{active adjustment of contact force}~\citep{housden2021towards}. The designed F/T sensor consists of two pieces with eight legs in total and the displacements of the legs were measured with eight optoelectronic sensors. 
By using the measured force as feedback, this system can actively adjust the contact force toward the desired values~\citep{housden2021towards}. Bao~\emph{et al.} designed a parallel, motor-spring-based end-effector to actively generate a certain force for US scanning~\citep{bao2021constant}. The force is adjusted by changing the position of two sliders connecting a moving platform using springs. The symmetrical configuration restricted the contact force consistent with the probe's centerline. 

\subsubsection{Others}
\par
Huang~\emph{et al.} attached two thin force sensors (IMS-Y-Z03, I-Motion Inc., China) on both sides of the front face of a linear probe~\citep{huang2018fully}. Then, a simple rule was implemented to control the applied force: the probe moves downward $3.1~mm$ when the force is smaller than $1~N$, the probe moves upward $3.1~mm$ when the force is larger than $8~N$, and scans were only performed when both sensors measurements are in the range of $[1, 8~N]$. Their team extended this work by replacing a 3-DOF linear stage with a 6-DOF robotic arm~\citep{huang2018robotic}. A robotic arm enables in-plane rotation; thereby, an updated rule was used to maintain the constant force: the probe moves downward $0.2~mm$ when both the forces are smaller than the desired force, the probe moves upward $0.2~mm$ when the forces are larger than the desired one, the probe rotates $0.2^{\circ}$ (in-plane) when the two forces are different. Compared with other force adjustment approaches, this method is easy to be implemented, while the handcraft rule needs further improvement to adapt to inter-patient variations. 



\subsection{Probe Orientation Optimization}~\label{sec_IV_B_enable_orj}
\par
The relative probe orientation with respect to the restricted surface is also a key factor dominating the image quality. For some applications like US imaging of bone, US probe orientation is often optimized to be orthogonal to the constraint surface~\citep{ihnatsenka2010ultrasound,jiang2020automatic_tie_normal}. 
\revision{In certain applications}, such as image-guided interventions, the US probe may need to be tilted from the orthogonal direction in order to better visualize the targets and/or \revision{inserted instruments}~\citep{zettinig2016toward}. The articles discussing probe orientation adjustment \secRevision{are} summarised in three subcategories: \textbf{in-plane orientation}, \textbf{out-of-plane orientation}, and \textbf{full orientation} optimization in this section.


\begin{table*}[!ht]
\normalsize 
\centering
\caption{\revision{Representative Probe Orientation Determination and Optimization Approaches used in RUSS}}
\label{Table_orientation}
\resizebox{\textwidth}{!}{
\revision{
\begin{tabular}{cccccc}
\noalign{\hrule height 1.2pt}
Probe orientation & Reference & Key signal & Robot & Applications & Data\\ 
\noalign{\hrule height 1.0 pt}
\multicolumn{1}{c}{\multirow{4}{*}{\makecell[c]{In-plane\\orientation}}}  & \cite{chatelain2015optimization,chatelain2016confidence}  &{}  & 6-DOF Viper s650 robot  &general  &phantom\\
\multicolumn{1}{c}{}   & \cite{jiang2022precise}   &US confidence map   &7-DOF KUKA iiwa  &vessel  &phantom\\
\multicolumn{1}{c}{}   & \cite{welleweerd2020automated}   &{}   &7-DOF KUKA iiwa  &breast  &phantom\\
\multicolumn{1}{c}{}   & \cite{zettinig2016toward}   &CT or MRI  &7-DOF KUKA iiwa  &spine  &phantom\\
\multicolumn{1}{c}{}   & \cite{huang2018robotic}   &two 1-DOF load cells   &6-DOF robotic arm   &general  &phantom\\
\noalign{\hrule height 0.5 pt}
\makecell[c]{Out-of-plane\\orientation} & \cite{virga2016automatic}  &US confidence map   & 7-DOF KUKA iiwa  &aorta  &volunteer\\
\noalign{\hrule height 0.5 pt}
\multicolumn{1}{c}{\multirow{6}{*}{\makecell[c]{Full orientation}}}  & \makecell[c]{\cite{jiang2021motion};\\\cite{ma2021novel};\\\cite{ma2021autonomous}}   &RGB-D images   & \makecell[c]{7-DOF KUKA iiwa\\6-DOF UR5e \\ 7-DOF Franka Pamda}  &\makecell[c]{vessel\\vessel \\ lung}  &phantom\\
\multicolumn{1}{c}{}   & \cite{jiang2020automatic} &force \& US images    &7-DOF KUKA iiwa   & general &phantom\& volunteer\\
\multicolumn{1}{c}{}   & \cite{jiang2020automatic_tie_normal}  &force   &7-DOF KUKA iiwa     & general (orthopedic) &phantom \& volunteer\\
\multicolumn{1}{c}{}   & \makecell[c]{\cite{welleweerd2021out};\\\cite{chatelain2017confidence}}  & US confidence map  &7-DOF KUKA iiwa   & \makecell[c]{breast\\general} & \makecell[c]{phantom\\phantom\&volunteer}\\
\multicolumn{1}{c}{}   & \cite{osburg2022generalized}  & 3D US images  &7-DOF KUKA iiwa   & general & phantom\\
\multicolumn{1}{c}{}   & \cite{jiang2021autonomous_TIE_vessel}  & 2D US images  &7-DOF KUKA iiwa   & vessel & phantom\\
\noalign{\hrule height 1.2 pt}
\end{tabular}
}
}
\end{table*}

\subsubsection{In-Plane Optimization} 
\par
The in-plane orientation of a 2D probe represents the rotation around the short axis of the probe \revision{(see Fig.~\ref{Fig_searchingpolicy})}. In other words, in-plane motion only happens in the plane of US view. 
In~\citep{chatelain2015optimization}, the in-plane rotation was optimized using the visual servoing technique to improve the general image quality. To quantitatively assess the image's quality and further use it as the input signal for servoing control, the US confidence map~\citep{karamalis2012ultrasound} was computed for individual images. The US confidence map provides a pixel-wise measure of signal loss based on a simplified model of wave propagation in tissues. 
\revision{The computed confidence map is often used as a measurement metric of image's quality~\citep{chatelain2015optimization}. However, it is worth noting that the quality here refers only to the strength of US signal. The best US images according to the confidence map may not be the best images expected by clinicians in examinations. 
To obtain the US images leading to higher overall confidence values,} the probe's orientation was often optimized to the orthogonal direction of the surface~\citep{chatelain2015optimization}. 
\revision{In addition, Jiang~\emph{et al.} and Welleweerd~\emph{et al.} also employed US confidence map-based in-plane adjustments to improve sub-optimal contact conditions for limb arm and breast scans~\citep{jiang2022precise, welleweerd2020automated}, respectively.}

\par
Huang~\emph{et al.} adjusted in-plane orientation to balance the contact forces measured at two endpoints on the probe tip~\citep{huang2018robotic}. 
Zettinig~\emph{et al.} proposed a 3D-to-3D volume registration to adapt the movement of target anatomy; then they further optimized the in-plane orientation to align the current needle guideline \revision{with} the planned path on a preoperative CT or MR~\citep{zettinig2016toward}.

\subsubsection{Out-of-Plane Optimization} 
\par
The out-of-plane motion is defined as the rotation around the probe's axial direction \revision{(see Fig.~\ref{Fig_searchingpolicy})}. 
\secRevision{In~\citep{virga2016automatic}, authors claimed that} in-plane adjustment \secRevision{only} benefit axial aortic scans marginally; therefore, they optimized out-of-plane rotation to improve the imaging quality in terms of overall US confidence values~\citep{virga2016automatic}. A fixed rotation angle interval was applied step by step. However, it is uncommon for existing articles to only optimize the out-of-plane orientation.

\subsubsection{Full Orientation Optimization} 
\par
\secRevision{To estimate the normal direction of a constrained surface}, depth camera-based approaches are most often used in the existing literature~\citep{ma2021novel, jiang2021motion, ma2021autonomous}. The advantage of these approaches is high computational efficiency, while the main limitation is \secRevision{relatively} low accuracy of the estimations.
\secRevision{
Recently, Ma~\emph{et al.} designed a probe holder with four laser distance sensors to actively adjust the probe's orientation to be normal to the surface~\citep{ma2022see}. The results demonstrated their adjustment can be computed in real-time.}
In addition, Jiang~\emph{et al} proposed a method to identify the normal direction of the restricted surface using contact force for out-of-plane optimization and US images for in-plane optimization~\citep{jiang2020automatic} (see~Fig.~\ref{Fig_searchingpolicy}). The bone boundary was used to demonstrate the probe orientation's impact on the imaging quality. In this study, Jiang~\emph{et al} proposed a feature called the smooth derivative of contact force, which enabled the accurate estimation of the out-of-plane orientation without the requirement for an expensive external F/T sensor~\citep{jiang2020automatic}. To further improve the accuracy of the estimated normal direction, Jiang~\emph{et al.} deduced the underlying mechanical model based on the force measured during two orthogonal fan motions at a given contact point~\citep{jiang2020automatic_tie_normal}. The upgraded method works for both convex and linear probes, and due to its purely force-based nature, it is invariant to \secRevision{image noises}. Yet, due to nonnegligible deformations of the soft tissue (e.g., breast), the force-based approaches are more suitable for orthopedic applications (e.g., limbs and back).

\par
Besides, a number of studies optimized the probe's full orientation solely using US images. Welleweerd~\emph{et al.} proposed a framework for automatic breast scanning without \revision{requiring patient-specific models}~\citep{welleweerd2021out}. To achieve this, in-plane optimization was firstly carried out to ensure acoustic coupling between the probe and \secRevision{the} examined breast. Once the mean confidence value~\citep{karamalis2012ultrasound} of the resulting image is inside the given range, the probe \revision{will be moved tangentially} to the breast. \revision{If the current mean confidence value differs from the specified range, out-of-plane corrections will be carried out to maintain constant confidence.}
The mean error between the estimated normal directions and ground truth at all points of trajectory was $12.6^{\circ}$ out-of-plane and $4.3^{\circ}$ in-plane~\citep{welleweerd2021out}. Chatelain~\emph{et al.} extended their preliminary work~\citep{chatelain2015optimization, chatelain2016confidence} from in-plane control of a 2D probe to full-orientation control of a 3D wobbler probe using the confidence map~\citep{chatelain2017confidence}. \revision{Recently, Osburg~\emph{et al.} used Convolutional Neural Network (CNN) to compute the surface normal at the point of contact based on native 3D volumetric data~\citep{osburg2022generalized}. 
}

\par
Instead of identifying the normal direction of \revision{constraint surfaces}, Jiang~\emph{et al.} \revision{estimated the normal direction of a subcutaneous tubular structure directly based on the segmented vessels of the most recent images}~\citep{jiang2021autonomous_TIE_vessel}. The vascular boundaries obtained at different positions contain the local geometrical information (radius and centerline) of the blood vessel; thus, the US probe can be oriented orthogonally to the estimated centerline of the local segment of the tubular structure.

\begin{table*}[!ht]
\normalsize 
\centering
\caption{\revision{Representative Scan Path Generation Approaches used in RUSS}}
\label{Table_path}
\resizebox{\textwidth}{!}{
\revision{
\begin{tabular}{cccccc}
\noalign{\hrule height 1.2pt}
Method & Reference & Key signal & Robot & Applications & Data\\ 
\noalign{\hrule height 1.0 pt}
\multicolumn{1}{c}{\multirow{18}{*}{\makecell[c]{Off-line\\path planning}}}  & \cite{merouche2015robotic}  &manual  & 6-DOF robotic arm  &vessel  &phantom\\
\multicolumn{1}{c}{}   & \cite{akbari2021robot}   &US images   &7-DOF Franka Pamda  &breast  &phantom\\
\multicolumn{1}{c}{}   & \makecell[c]{\cite{jiang2022towards};\\\cite{langsch2019robotic};\\\cite{virga2016automatic};\\\cite{graumann2016robotic};\\\cite{hennersperger2016towards}}  &\makecell[c]{CT/MRI\\\&RGB-D images}   &7-DOF KUKA iiwa  &vessel  &\makecell[c]{volunteer\\phantom\&volunteer\\volunteer\\phantom\\volunteer}\\
\multicolumn{1}{c}{}   & \cite{mustafa2013development}   &RGB images &6-DOF robotic arm  &liver  &volunteer\\
\multicolumn{1}{c}{}   & \cite{jiang2022precise}   &{}   &7-DOF KUKA iiwa   &vessel  &phantom\\
\multicolumn{1}{c}{}   & \cite{huang2018fully,huang2018robotic}   &{}   &6-DOF robotic arm   &general  &phantom\\
\multicolumn{1}{c}{}   & \cite{ma2021novel}   &{}   &6-DOF UR5e   &vessel  &phantom\\
\multicolumn{1}{c}{}   & \cite{suligoj2021robust}   &RGB-D images  &7-DOF Franka Pamda   &vessel  &phantom\\
\multicolumn{1}{c}{}   & \cite{ma2021autonomous}   &{}   &7-DOF Franka Pamda   &lung  &phantom\\
\multicolumn{1}{c}{}   & \cite{wang2022full}   &{}   &6-DOF UR5   &breast  &phantom\\
\multicolumn{1}{c}{}   & \cite{tan2022fully,tan2022automatic}   &{}   &customized mechanism    &lung/breast  &volunteer\\
\multicolumn{1}{c}{}   & \cite{gobl2017acoustic}   &\makecell[c]{CT/MRI\\\&acoustic information}    &7-DOF KUKA iiwa  &\makecell[c]{thorax\\(liver/heart)}  &phantom\\
\multicolumn{1}{c}{}   & \cite{sutedjo2022acoustic}   &\makecell[c]{3D US volume\\\&acoustic information}   &7-DOF KUKA iiwa  &thorax   &phantom\\
\noalign{\hrule height 0.5 pt}
\multicolumn{1}{c}{\multirow{5}{*}{\makecell[c]{On-line\\ path planning}}}  & \makecell[c]{\cite{jiang2021autonomous_TIE_vessel}}   &{}   & 7-DOF KUKA iiwa &vessel  &phantom\\
\multicolumn{1}{c}{}   & \cite{huang2021towards} &2D US images    &6-DOF UR5   & carotid &volunteer\\
\multicolumn{1}{c}{}   & \cite{kim2020robot}  &{}   &7-DOF Kinova Gen2     & cardiac &phantom \\
\multicolumn{1}{c}{}   & \makecell[c]{\cite{welleweerd2021out}}  & US confidence map  &7-DOF KUKA iiwa   & breast & phantom\\
\multicolumn{1}{c}{}   & \makecell[c]{\cite{jiang2023dopus}}  & US \& Doppler images  &7-DOF KUKA iiwa   & vessel & volunteer\\
\noalign{\hrule height 1.2 pt}
\end{tabular}
}
}
\end{table*}

\subsection{Path Generation for Autonomous US Scanning}~\label{sec_IV_C_enable_path}
\par
\revision{In order to accomplish US examinations, a \secRevision{proper path is essential} to visualize the object or locate the lesion on human tissue,} e.g., along a target blood vessel and covering a volume of interest. This section categorizes the existing path planning methods as 1) offline scan path generation methods and 2) online scan path generation methods.


\begin{figure*}[ht!]
\centering
\includegraphics[width=0.95\textwidth]{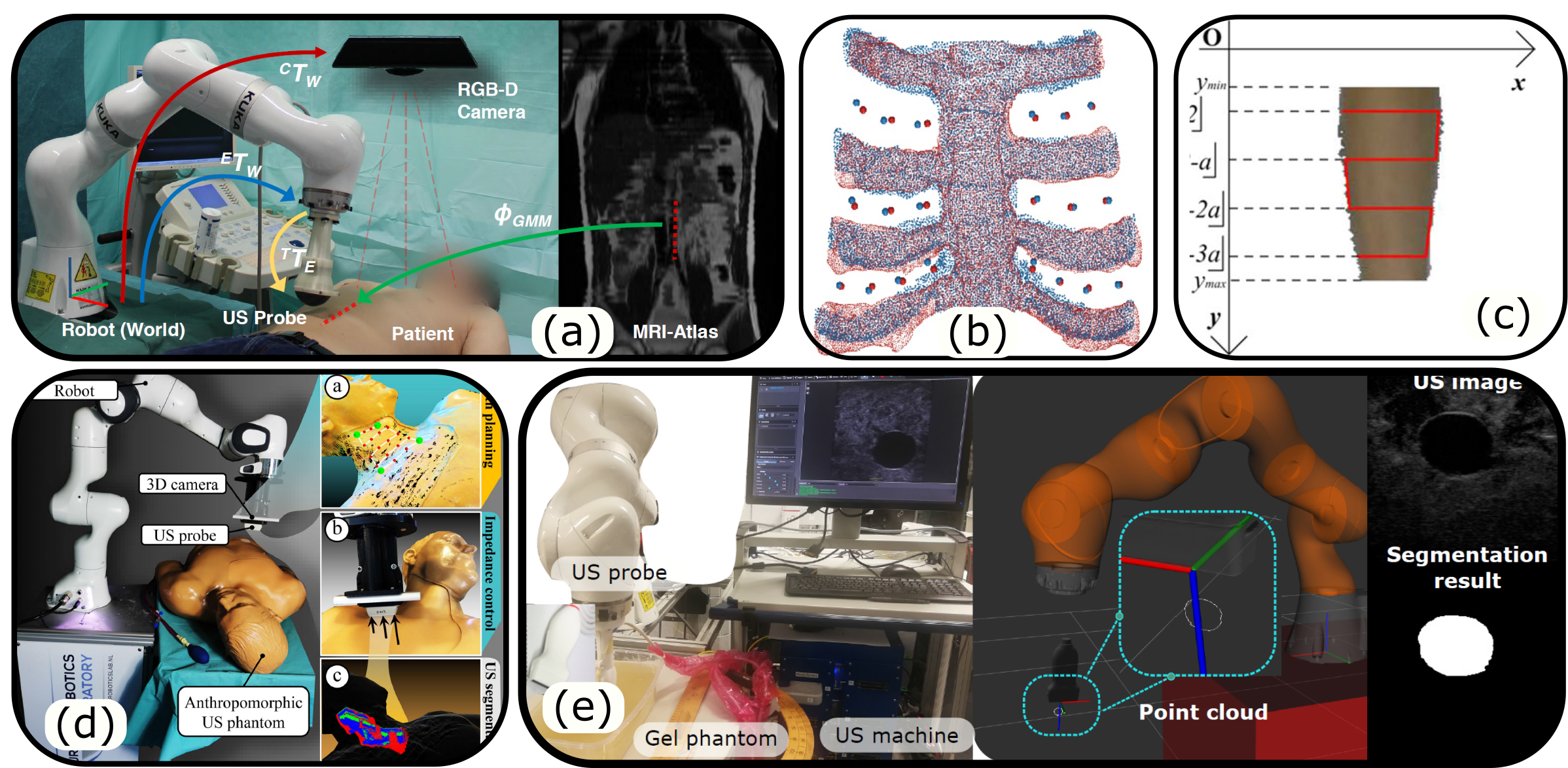}
\caption{Representative approaches of scanning path panning. (a) Manually plan a scanning trajectory on preoperative MRI image and further transfer the planning path to the robot frame by performing the registration between the MRI data and the 3D RGB-D surface image acquired using a Kinect camera~\citep{virga2016automatic}. (b) scan trajectory mapped from CT to US spaces~\citep{jiang2023thoracic, jiang2023skeleton}. (c) and (d) automatically generate a ``snake" trajectory on the extracted target surface from RGB images~\citep{huang2018fully} and~\citep{suligoj2021robust}, respectively. (e) online path planning approach using real-time US image feedback to automatically perform screening along tubular tissues~\citep{jiang2021autonomous_TIE_vessel}. 
}
\label{Fig_path_planning}
\end{figure*}

\subsubsection{Offline Scan Path Generation}
\par
To locate and evaluate the length and severity of stenosis for planning the treatment of peripheral arterial disease (PAD), Merouche~\emph{et al.} directly give the scanning path by manually moving the robotic arm along the target artery~\citep{merouche2015robotic}. To address the potential visualization issue caused by small motions after path planning procedures and to facilitate the tracking of the artery during automatic scans, the probe's position was tuned to maintain the cross-sectional lumen horizontally centered in the US view. Similarly, Jiang~\emph{et al.} \revision{manually drew a scan path on the surface of a vascular phantom, and then extracted the path based on RGB images}~\citep{jiang2021motion}.

\par
\secRevision{Considering autonomous path planning,} scan trajectories can be determined on pre-scanned images (e.g., MRI and CT); then, transferring the planned path to the current setup by registering the live US or RGB-D image to the preoperative atlas.
Hennersperger~\emph{et al.} validated the feasibility of autonomously transferring a planned scan path from MRI to the current setup based on the registration between the MRI and 3D \secRevision{surface point clouds} acquired by a Kinect camera (Microsoft Corporation, USA)~\citep{hennersperger2016towards}. Similarly, Langsch~\emph{et al.} computed the scanning trajectory of an aorta by registering 3D US volume to the patient's MRI~\citep{langsch2019robotic}. However, due to the need for tomographic data (MRI or CT) of each patient, the advantage of these approaches is reduced in clinical practice. To further address this challenge, Virga~\emph{et al.} carried out non-rigid registration between the patient-specific 3D surface extract from a depth camera and a generic preoperative MRI template~\citep{virga2016automatic} [see Fig.~\ref{Fig_path_planning} (a)].
\final{Specific to thorax examinations, Jiang~\emph{et al.} presented a skeleton graph-based non-rigid registration between the cartilage point clouds extracted from a tomographic template and US images of patients~\citep{jiang2023skeleton}. To further improve the registration accuracy, Jiang~\emph{et al.} introduced the dense skeleton graph to replace the manually designed key points of the skeleton~\citep{jiang2023thoracic} [see Fig.~\ref{Fig_path_planning} (b)].} 
Akbari~\emph{et al.} presented a complete US-based approach to find a proper trajectory for breast US imaging~\citep{akbari2021robot}. A manual prior scan is carried out in advance; then, the desired trajectory for the post scan is computed based on geometrical analysis of the target using the pre-scanned US images.

\par
In addition, the scanning path is often planned solely on the surface extracted by an external camera directly~\citep{huang2023motion}. Mustafa~\emph{et al.} extracted the patient's abdomen surface from an RGB image acquired using a web camera (2D) based on a preset HSV color filter; then, the position of the liver was estimated and a four-step acquisition protocol was applied~\citep{mustafa2013development}. Due to the lack of imaging depth information, the camera needed to be carefully configured anteriorly to subjects. Ma~\emph{et al.} used a Realsense SR305 RGB-D camera (Intel Corporation, USA) to extract the 3D surface data using a depth threshold and further planned the scanning path on the extracted 3D surface~\citep{ma2021novel}. 
Huang~\emph{et al.} extracted 2D skin surfaces of patients from an RGB image using the rule ``red$>$Green$>$Blue"~\citep{huang2018fully, huang2018robotic} [see Fig.~\ref{Fig_path_planning} (c)]. They claimed this is more generic and robust than the threshold-based approaches. Then, a ``snake" trajectory was automatically generated to cover the area of interest. Suligoj~\emph{et al.} used the same logic to generate scan paths over a region manually \secRevision{annotated} in an RGB image~\citep{suligoj2021robust} [see Fig.~\ref{Fig_path_planning} (d)]. Recently, Ma~\emph{et al.} proposed a learning-based method to extract the human abdomen from a depth camera, and further divided the extracted region into four parts for autonomously generating scanning paths of \revision{the lung}~\citep{ma2021autonomous}. 


\par
The aforementioned path planning approaches for US scanning were directly determined on the patient's surface. However, the optimal coverage of an underlying volume of interest is not considered. To address this challenge, Graumann~\emph{et al.} proposed a method to automatically compute a suitable scanning path to cover a volume of interest easily selected in preoperative images~\citep{graumann2016robotic}. \revision{Depending on the sizes of targeting volumes}, one or multiple lines were automatically generated for full coverage. \revision{To automatically determine the optimal probe position on the skin to monitor the motion of the internal organ of interest, Bruder~\emph{et al.} computed patient-specific US image quality from a given CT scan~\citep{bruder2011d, bruder2014method}.}
To further consider the full coverage of subcostal organs like liver and heart, G\"obl~\emph{et al.} proposed a framework integrating both geometrical and physics-based constraints to estimate the best US scanning path with respect to the limited acoustic windows~\citep{gobl2017acoustic}. The poses maximizing the image quality (i.e., less acoustic attenuation) are finally selected. The results on both human and phantom data demonstrated that superior image quality was achieved using their method in comparison with a naive planning approach while maintaining the necessary coverage of the target. 



\subsubsection{Online Scan Path Generation}
\par
Although the off-line path planning are more often used in RUSS, some online planning approaches \revision{based on live US images have also been developed}. Online approaches can generate more flexible trajectories than offline approaches, which can effectively guarantee the target's visibility inside the US view, even in the presence of unexpected motion. In~\citep{jiang2021autonomous_TIE_vessel}, Jiang~\emph{et al.} proposed a pipeline to enable a RUSS to automatically perform US screening of tubular structures based only on real-time US image feedback. The US probe was manually positioned on the tubular structures [see Fig.~\ref{Fig_path_planning} (e)]. Afterward, a U-Net was activated to constantly segment cross-sectional vessel lumen from US images; and thereby, a set of boundary point clouds were extracted and further used to estimate the geometry (centerline and radius) of the local artery sections. To completely scan the whole artery, the US probe was moved forward in the direction of the estimated local vessel centerline in real-time. In addition, similar work was accomplished by Huang~\emph{et al.} for automatically screening of carotid \revision{artery} based on the US image feedback~\citep{huang2021towards}. In~\citep{kim2020robot}, Kim~\emph{et al.} employed a CNN as a classifier for real-time B-mode images to update the probe position for heart examinations. Since the next action is planned in real-time, the online path planning approach can facilitate the robust tracking of the target during autonomous scans. \final{To ensure the scanning quality to facilitate the clinical diagnosis, Jiang~\emph{et al.} first presented an online segmentation quality-aware method based on the Doppler signal~\citep{jiang2023dopus}. Once the segmentation performance is considered low, the probe orientation will be adjusted to enhance the Doppler signal and thereby improve the accuracy and completeness of the reconstructed 3D vessel. The significance of this study lies in its ability to inspire future research into quality-aware, closed-loop robotic scanning. 
}




\section{Application-Oriented \\Advanced Technologies for Autonomous RUSS}~\label{sec_V_Adv}
\par
The aforementioned three enabling technologies (force control, orientation optimization, and scanning path generation) have been \revision{extensively studied} in the existing literature. 
However, the enabling technologies can only guarantee the quality of US acquisition in ideal cases. To further enable the implementation of extensive and autonomous RUSS screening programs, more advanced technologies tackling practical challenges in real scenarios should be considered. In this section, four distinctive techniques are discussed: 1) \textbf{Motion-aware US imaging}: regarding the autonomous scanning of the anatomy of interest, the potential body motion should be monitored and properly compensated to achieve accurate and complete 3D anatomy geometry. 2) \textbf{Deformation-aware US imaging}: due to the inherited characteristic of US imaging, a certain force is necessary for properly visualizing the underlying anatomy of interest; thereby, the inevitable force-induced deformation hinders the correct measurements of the target anatomy. 3) \textbf{US visual servoing}: by providing pixel-to-pixel control to accurately move the probe to reach the desired cross-sectional images and guarantee the visibility of the object of interest in US views. 4) \textbf{Elastography imaging}: benefiting from the accurate control over probe position and contact force between the probe and tested objects, the underlying tissue properties can be estimated for diagnosis using RUSS. 


\subsection{Motion-Aware US Imaging}~\label{sec_V_A_Adv_motion}

\begin{figure}[ht!]
\centering
\includegraphics[width=0.40\textwidth]{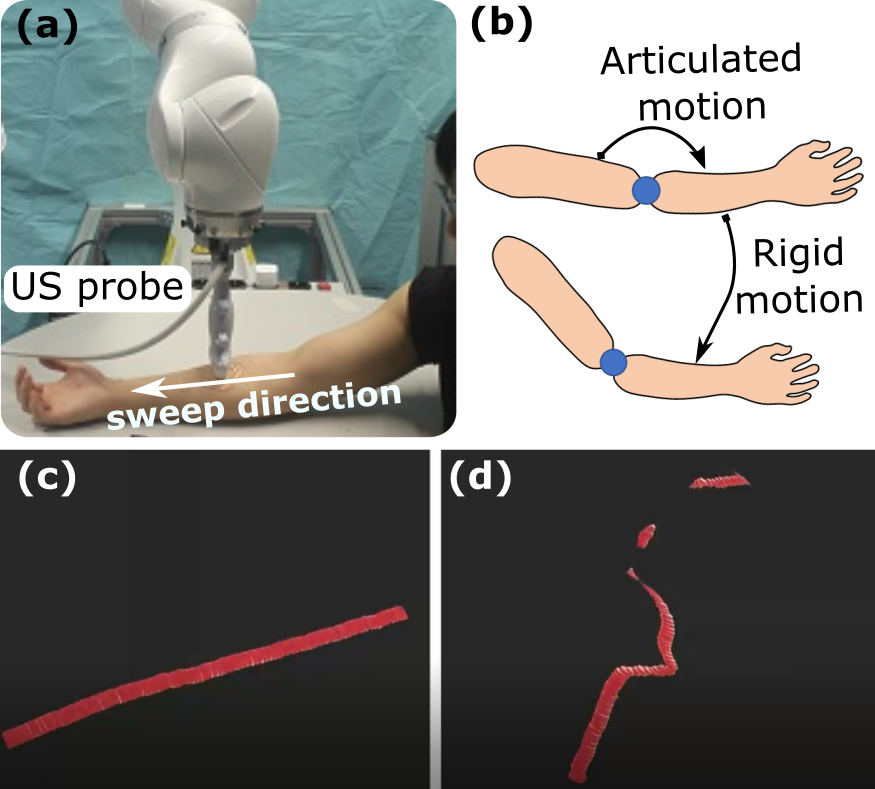}
\caption{Illustration of the 3D compounding results influenced by the object motion. (a) robotic US scanning on a volunteer's arm, (b) two representative types of motion: articulated motion and rigid motion, (c) 3D reconstruction result of the vessel when the imaged object is stationary during the sweep, and (d) 3D reconstruction result of the same vessel when the target is moved during the sweep. 
}
\label{Fig_shift}
\end{figure}


\begin{figure*}[ht!]
\centering
\includegraphics[width=0.90\textwidth]{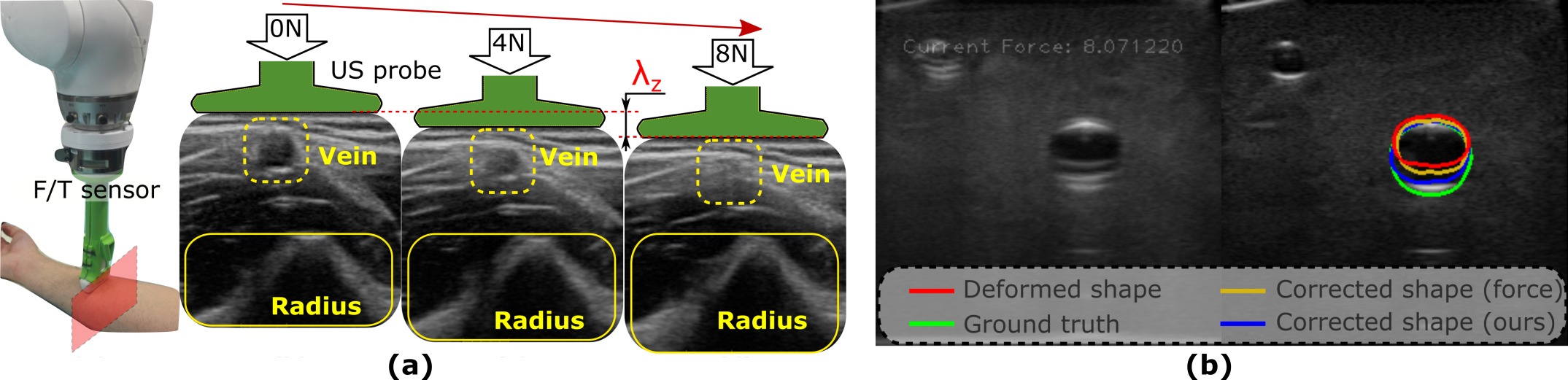}
\caption{ (a) Illustration of the force-induced anatomy (cephalic vein) deformation, and (b) a representative deformation correction result~\citep{jiang2021deformation}. 
}
\label{Fig_defromation_background}
\end{figure*}

\subsubsection{\secRevision{Periodic Motion Detection and Compensation}}
\par
In this context, periodic or quasiperiodic motions refer primarily to \secRevision{internal physiological motions such as respiration and pulsation.} Because of the advantages of non-invasive and real-time performance, US can be used to monitor internal tissue motion~\citep{fast2016first, ipsen2018visual, wulff2019robust}. In free-hand mode, it is extremely difficult to compensate \secRevision{for such motions to achieve} stable US images. \secRevision{To tackle this challenge}, RUSS has been seen as a promising solution~\citep{ipsen2021towards} because \secRevision{robots usually can provide higher accuracy in terms of positioning and repeatability than humans~\citep{haidegger2010importance}.} Esteban~\emph{et al.} reported that RUSS can intrinsically compensate for small motions caused by breathing or human tremor using compliant force control~\citep{esteban2018robotic}. Heunis~\emph{et al.} employed a 6-DOF Stewart platform to mimic the involuntary periodic movements that occur during scans; and further proposed a pipeline to create an effective scanning path to cover a surface while compensating for these motions and adhering to preset contact forces~\citep{heunis2021real}. This movement was also compensated for by using force control. The results demonstrated that the reconstruction error of arteries was $1.9\pm0.3~mm$ in non-static scenarios. To actively compensate for the respiration-induced motion in the liver or prostate, Ipsen~\emph{et al.} applied a constant force control to accomplish continuous US scans in long-term monitoring~\citep{ipsen2021towards}. Furthermore, visual servoing (\textbf{Section~\ref{sec_V_C_Adv_servo}}) is another \secRevision{potential} solution for compensating the respiration motion~\citep{nadeau2011automatic} and pulsation caused by heart beating~\citep{nadeau2014intensity}.

\subsubsection{\secRevision{Non-Periodic Motion Detection and Compensation}}
\par
Subjects are often adjusted by sonographers to better visualize the target during scans. Thus, the ability to compensate for non-periodic patient’s motion is crucial for the practical use of RUSS. A representative example of the influence caused by non-periodic motion of the imaged patients is shown in Fig.~\ref{Fig_shift}. The scanned results are significantly different when the same object is kept stationary and moved during \secRevision{scanning}.

\par
To obtain complete and accurate 3D US scans of a vascular phantom in the presence of rigid motion, Jiang~\emph{et al.} proposed a vision-based RUSS to actively compensate for such non-periodic motion~\citep{jiang2021motion}. In this study, five passive markers were rigidly attached to the imaged phantom surface and further used to monitor the potential target motion. Once the target is moved, the motion-aware RUSS automatically computes the transformation and updates the trajectory to recover the scanning from the breaking point. \revision{To eliminate the requirement for careful configuration of the passive markers in real scenarios, Jiang~\emph{et al} monitored the patient's motion based on the real-time segmentation of objects in RGB images and computed the compensation matrix using extracted surface point clouds acquired before and after the motion~\citep{jiang2022precise}. The results on a \revision{realistic} arm phantom demonstrate the effectiveness of this marker-less compensation method. The advantages of robotic US (accuracy and stability) and free-hand US (flexibility) were combined by including active compensation for potential patient motion during scans.} However, such systems only considered the rigid motion of objects. To further tackle \secRevision{non-rigid} articulated joint motions, Jiang~\emph{et al.} proposed a vision-based framework, combining joint detection and non-rigid surface registration, to automatically update scanning trajectories from a template to \secRevision{individual volunteers with varying arm gestures}~\citep{jiang2022towards}. The robustness and accuracy of the proposed system have been evaluated on multiple volunteers. 




\subsection{Deformation-Aware US Imaging}~\label{sec_V_B_Adv_def}

\par
\revision{
Due to the probe-patient contact force, shape distortion of the visualized anatomy's geometry is inevitable, particularly for soft tissues such as superficial blood vessels (see Fig.~\ref{Fig_defromation_background}). The force-induced deformation reduces the precision and repeatability of US images, and thereby could further limiting the diagnostic accuracy and consistency, especially for computer-assisted diagnosis.}


\par
\revision{To provide precise and reliable US images, pressure-induced image deformation needs to be properly corrected. Unlike human sonographers, robots/computers are not trained to make the diagnosis based on deformed images. Therefore, such corrections are particularly important for RUSS.}
To achieve distortion-free images, Treece~\emph{et al.} combined non-rigid image-based registration with position sensing to correct pressure-induced deformations for free-hand 3D imaging~\citep{treece2002correction}. Sun~\emph{et al.} computed 2D deformation fields based on the estimated pixel displacements and corresponding contact forces using polynomial regression models~\citep{sun2010trajectory}. The pixel displacements were computed based on flow techniques using raw echo frequency (RF) data. Based on their experimental results, the parabolic polynomial regression model significantly outperforms the linear model. However, there was no significant performance difference between $2nd$ order and higher-order polynomial models. Burcher~\emph{et al.} build a model using the finite element method (FEM) to predict the deformation~\citep{burcher2001deformation}. Nonetheless, the performance of the FEM-based approach is heavily dependent on the prior knowledge of tissue properties, which are usually hard to measure in real scenarios. To overcome this challenge, Dahmani~\emph{et al.} employed a linear elastic model to approximate personalized biomedical properties of involved tissues from the images~\citep{dahmani2017model}.

\par
\revision{To alleviate the inter-variation of pressure-induced deformation between the acquired images along a scanning path, RUSS is often required to maintain a constant force during the screening.} 
To \secRevision{correct distorted} images, Virga~\emph{et al.} built a 4th-order polynomial model to regress the pixel displacement with respect to contact force and further propagate the computed deformation field at sparse sampling points to the whole sweep direction~\citep{virga2018use}. The sampling points were selected manually on the first frame and this method took $186~s$ on average to compute a deformation field at one location. To speed up the process for compression-free 3D volume, Jiang~\emph{et al.} proposed a stiffness-based deformation correction approach, incorporating image pixel displacements, contact forces, and nonlinear tissue stiffness~\citep{jiang2021deformation}. To obtain patient-specific stiffness models, robotic palpation was performed at sampling positions. Since tissue stiffness is the key factor dominating the deformation, the optimal deformation regression models at sampling positions can be propagated to other positions on the trajectory by interpolating the estimated local stiffness. \secRevision{However, the state of the art in the field of US image correction for force-induced deformation is not yet applicable to clinical practice.} To further achieve this objective, a pixel-wise tissue properties estimator and anatomy-aware correction system should be developed to bridge the gap between different anatomy and different patients.

\subsection{Ultrasound Visual Servoing}~\label{sec_V_C_Adv_servo}
\par
Understanding the interaction of sonographers with the patient and the US probe is of high importance when developing RUSS. In order to acquire B-mode images of the anatomy of interest, sonographers perform a rough positioning of the probe on the human body. Consecutively, the B-mode images are analyzed while adjusting the probe to obtain the final view with the anatomy of interest in focus. This dynamic image-based adjustment and exploring of the anatomy can be defined as ``visual servoing". While this has been the subject of research in the last decades, we believe that the introduction of deep learning and the advances in reinforcement learning could allow the scientific community to further understand and solve this image-based optimization problem. Recent work that has been published in this field~\citep{bi2022vesnet, hase2020ultrasound, li2021autonomous} can be taken as an indicator for being a potentially interesting research topic in the coming years. In this section, we review some prior work on visual servoing that can be considered as a development of the state of the art towards the goal of autonomous intelligent exploration of particular anatomy and physiology views needed for examination and treatment.

\subsubsection{Autonomous US Probe Guidance}
To automatically rediscover a previously registered US imaging view, Bachta~\emph{et al.} developed an image-based visual servoing approach using boundary information and tested \revision{it} in a simulator~\citep{bachta2006towards}. The target edge was retrieved using  a polynomial regression analysis, and the optimized coefficients were used as visual features to guide a robot-controlled probe to reach a desired image section. However, this method suffers from image noise and is limited to a specific shape. To overcome this challenge, Mebarki~\emph{et al.} employed image moments as visual features~\citep{mebarki2008image,mebarki2008automatic}, which are generic and robust with respect to measurement perturbations. To further achieve a model-free servoing task on unknown targets, they compute the interaction matrix in real-time using B-mode images~\citep{mebarki2009modeling,mebarki20102}. The experiments on gelatin phantoms demonstrated promising results in terms of minimizing the visual-features error; however, only local convergence can be guaranteed. In particular, in the case of a roughly symmetric object, similar geometric properties can be observed from different cross-sectional images. To overcome this shortage, Nadeau~\emph{et al.} defined a set of 2D features based on a three-dimensional space using a motorized 3D probe~\citep{nadeau2010multi, nadeau2016moments}.


\par
To accurately and \revision{actively navigate} the probe to a given US plane using the visual servoing technique, Duflot~\emph{et al.} first used the subsampled shearlet coefficients as novel visual features as an input to the controller, instead of pure image signal information, i.e., point, lines, moments, etc.~\citep{duflot2016towards}. Since a set of noiseless and redundant features can be extracted using shearlet coefficients, promising performances of their approach in terms of accuracy, repeatability, and robustness could be achieved. A comprehensive comparison between shearlet-based and photometric-based visual servoing controllers was carried out in both simulator and physical phantom~\citep{duflot2016shearlet, duflot2019wavelet}.

\par
\subsubsection{\secRevision{Imaging Stabilization and Object Tracking}}
\par
Visual servoing has also been used to track anatomies of interest and perform online compensation of the anatomy’s motion to stabilize the real-time US images. \secRevision{Without compensating for some potential motion like breathing, the resulting images will be affected. This will lead to inaccuracies in the estimation of the precise location of intervention target tissues. US visual servoing technologies are developed to compute the corresponding probe adjustment against environment dynamics based on real-time image feedback.}
Nadeau~\emph{et al.} presented an intensity-based approach to maintain the view of an organ while compensating for the physiological motion of the patient~\citep{nadeau2011improving}. Since the computation of image moments depends on object segmentation, image intensity values were directly used as visual features. In an extension work, they adapted their method for 3D probes and did first validations on soft animal tissues~\citep{nadeau2013intensity}. In 2015, Nadeau~\emph{et al.} applied a similar intensity-based visual servoing method to keep a target centered within a virtual imaging view in the context of intracardiac surgery~\citep{nadeau2014intensity}. Its effectiveness \secRevision{has been validated on in-vivo data}. Besides cardiac applications, Nadeau~\emph{et al.} applied visual servoing to stabilize respiratory motion by compensating periodic disturbances with a predictive controller~\citep{nadeau2011automatic}.

\par
In addition to intensity-based approaches, Krupa~\emph{et al.} employed US speckle information to estimate both in-plane and out-of-plane motion, thereby, realizing the tracking of soft tissue movements in US view~\citep{krupa2007real}. Speckle is often considered \revision{to be noise}, however, it conveys valuable data on the tissue of interest.
Speckle contains spatially coherent \revision{information} between consecutive US images \revision{because it physically results from coherent reflections of small components in human tissue.}
The preliminary experiments performed on a phantom with 2-DOF in-plane and out-of-plane motions demonstrated the potential of a speckle-based servoing approach. The validation for 6-DOF motion was further reported in~\citep{krupa2009real}. To further consider soft tissues' deformation, Royer~\emph{et al.} developed a physics-based model to facilitate the accurate tracking of the target of interest in 3D US images~\citep{royer2015real, royer2017real}.

\par
\subsubsection{Imaging Quality Optimization}
\par
Visual servoing techniques have also been investigated to improve imaging quality. Chatelain~\emph{et al.} first introduced the US confidence map as a new feature for visual servoing~\citep{chatelain2015optimization}. The authors claimed that the US imaging quality could be improved by optimizing the probe orientation to maximize the overall confidence value. An interesting extension using 3D probes instead of 2D probes has been reported in~\citep{chatelain2017confidence}. To evaluate the effect of the proposed method in real scenarios, in-vivo validations were performed on healthy volunteers. In addition, Patlan~\emph{et al.} directly employed elastography as the input of the visual servoing controller~\citep{patlan2016automatic}. To optimize the quality of the resulting elastography, the probe was automatically actuated to image a soft tissue object from different views, and further fused to enhance the \secRevision{computed} elastography. 



\subsection{Elastography Imaging}~\label{sec_V_D_Adv_Elas}
\par
US elastography is a non-invasive technique aiming to estimate the mechanical proprieties (i.e., stiffness) of the underlying soft tissues. Elastography has gained great interest in applications such as \revision{differentiating} tumors from healthy tissues (breast, prostate, liver, etc.) and guiding radiofrequency ablation surgeries~\citep{sigrist2017ultrasound}. Based on the underlying principles for producing US elastography, the currently available techniques can be mainly grouped into shear wave imaging and mechanical strain imaging. In shear wave imaging, the propagation speed of shear wave is measured. In addition, for strain imaging, a mechanical compression is performed using a US probe on the object's skin, where the mechanical compression process can be accurately controlled and measured based on robotic techniques. Thereby, accurate and standardized elastography is expected to be achieved. 

\begin{figure}[ht!]
\centering
\includegraphics[width=0.45\textwidth]{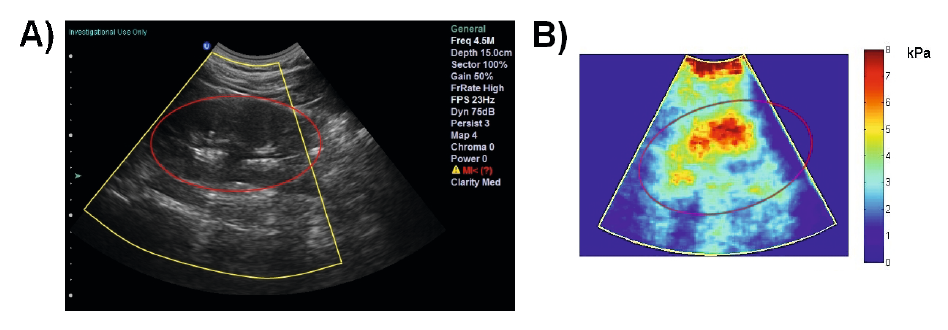}
\caption{Images of a healthy kidney circled in red. Left: B-Mode image. Right: Elastogram~\citep{schneider2012remote}. 
}
\label{Fig_elastography}
\end{figure}

\par
Compared with shear wave imaging, strain images are more common for robotic elastography imaging because it doesn't require specialized US hardware. Schneider~\emph{et al.} computed laparoscopic US elastography using an external vibrator positioned on the patient skin, where the US probe was remotely controlled by da Vinci (see Fig.~\ref{Fig_elastography})~\citep{schneider2012remote}. Patlan-Rosales~\emph{et al.} computed strain images using real-time radio-frequency (RF) signals to precisely locate subcutaneous tumors~\citep{patlan2016automatic}. In this study, robot-assisted palpation was used instead of an external vibrator and the resulting strain images were used to horizontally maintain the object in the imaging center. To estimate the strain map of moving tissues, Patlan-Rosales~\emph{et al.} estimated and compensated the non-rigid motion using visual servoing on an abdominal phantom~\citep{patlan2017strain}. Instead of 2D elastography, the same team extended their work to create 3D elastography based on the pre- and post-compressed volumes obtained by a 3D US probe~\citep{patlan2017robotic}.

\par
\revision{To compute 3D elastography without using a 3D probe,} Huang~\emph{et al.} designed a linear sliding track with a position sensor and a height-adjustable holder for conventional 2D probes~\citep{huang2015correspondence}. In this study, the pre- and post-compression echo signals were recorded by manually adjusting the height of the probe holder. Then, paired frames of RF data from the pre- and post-compression sweeps were obtained by interpolation. 2D strain images were computed using the paired RF data; thereby, 3D strain maps \secRevision{were} obtained by stacking the computed 2D strain images. To allow automatic acquisition of 3D strain maps, they \secRevision{replaced} the linear track with a motorized 3-DOF linear stage~\citep{chen2015development} and a 6-DOF robotic arm~\citep{yao2019three}, respectively.

\section{AI-Powered Robotic US Acquisition}~\label{sec_VI_AI}
\par
\revision{AI techniques have been seen as a promising way to further improve the automation level of RUSS by enhancing the understanding of US images and enabling the intuitive transfer of senior sonographers' advanced physiological knowledge. Such techniques have gained increasing attention most recently.} A diverse set of tasks like segmentation and classification of US images have achieved great success. Regarding the field of US image segmentation and classification, a large number of research articles have been published. More detailed techniques can be found in these survey articles~\citep{bass2021ultrasound, hesamian2019deep, shan2021ultrasound}. In this article, we will only focus on the studies that aim to automatize and/or standardize US scanning using AI-based approaches. More specifically, the approaches tried to automatically search for specific anatomical features or navigate a probe to display  standard US planes needed for examinations. These tasks are challenging \revision{because RUSS must be able to properly interpret the current states (US image, contact force, probe pose) and the surrounding context.}

\begin{figure*}[ht!]
\centering
\includegraphics[width=0.90\textwidth]{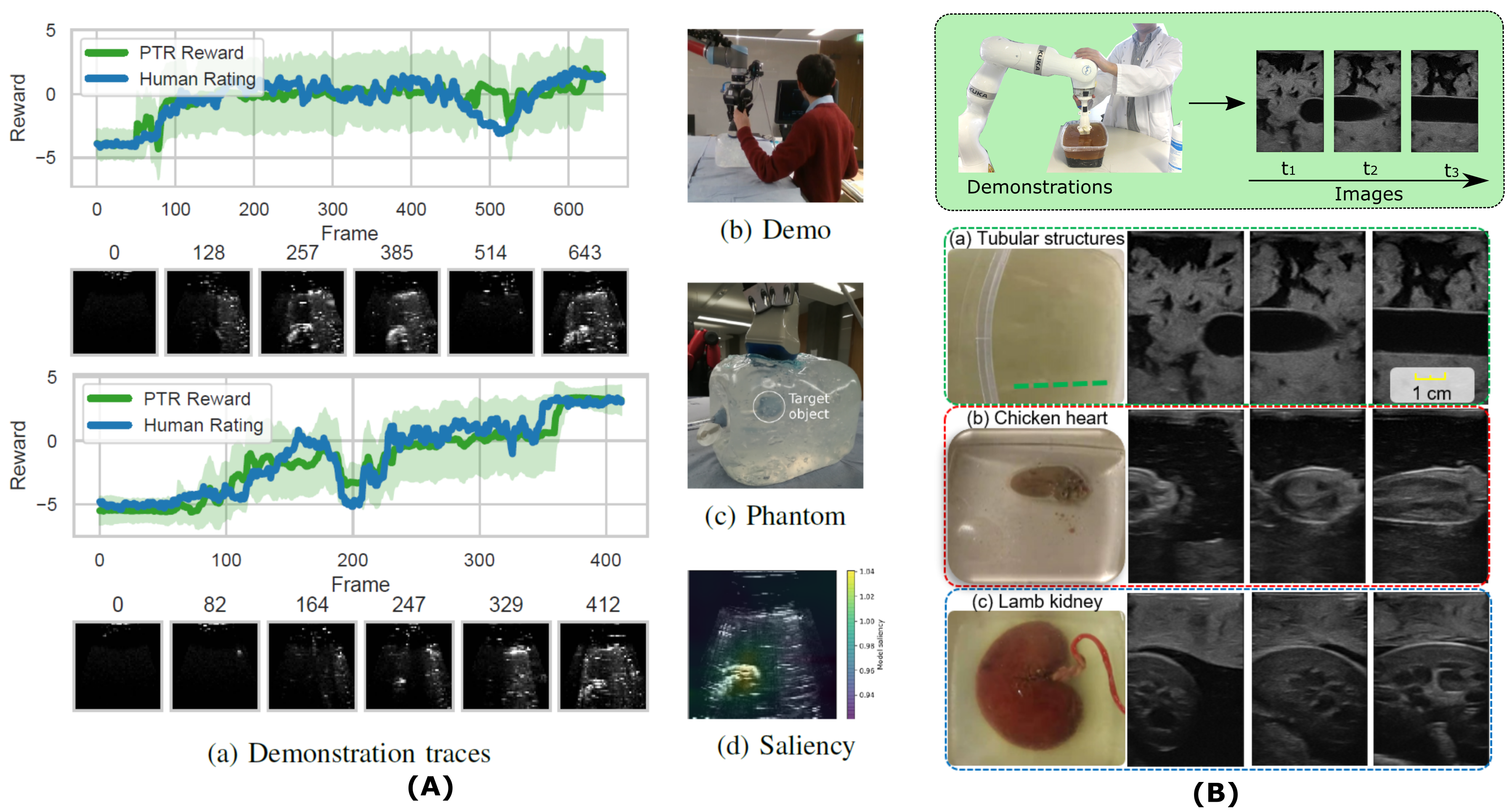}
\caption{(A) Learning from the expert demonstrations to automatically search for target anatomies using Probabilistic Temporal Ranking~\citep{burke2020learning}. (B) learning from demonstrations using Mutual Information-based Global Probabilistic Spatial Ranking~\citep{jiang2023intelligent}.
}
\label{Fig_AI_based_RUSS}
\end{figure*}

\par
Due to the potential tissue deformation and inconsistent acoustic artifacts of medical US \secRevision{images}, guiding a probe to visualize target objects in desired planes is a highly sophisticated task, which requires years of training~\citep{maraci2014searching}. However, such knowledge is not yet available for robots or computers. Due to the great advantage in feature representation over naive handcrafted features, CNN has the potential to achieve superhuman performance to robustly and accurately locate standard planes on challenging US images. Chen~\emph{et al.} employed a deep CNN to identify the fetal abdominal standard plane from recorded US video~\citep{chen2015standard}. Since data collection and manual labeling are time-consuming, a transfer learning strategy was used to guarantee the performance with limited training data. To achieve real-time performance, Baumgartner~\emph{et al.} proposed a deep CNN architecture called SonoNet to automatically detect $13$ fetal standard planes as well as provide localization of the fetal structures using a bounding box~\citep{baumgartner2017sononet}. The SonoNet was trained in a weakly supervised mode with only image-level scan plane labels, which make it possible to prepare a large data set. These approaches aid sonographers to locate standard planes that can also improve efficiency in particular for novices. Yet, these methods cannot automatically guide the probe towards target planes or anatomical structures of interest.


\par
To enable the ability of RUSS to automatically perform US scans, Mylonas~\emph{et al.} proposed a learning-based approach allowing autonomous execution of US scanning according to expert demonstrations~\citep{mylonas2013autonomous}. To achieve this objective, a Gaussian Mixture Modeling (GMM) was employed to model the demonstrations (trajectories) towards target objects in a probabilistic manner. However, since the real-time US image was not taken into consideration, all the demonstrations roughly started from the same initial position. This limitation severely impairs the usability of this method in real scenarios. To overcome this limitation and further provide real-time probe movement guidance for obtaining standard planes, Droste~\emph{et al.} proposed a behavioral cloning framework to mimic the process of sonographers searching for standard planes~\citep{droste2020automatic}. The proposed US-GuideNet consists of two fully connected layers and a gated recurrent unit (GRU) used to extract the sequential information. \revision{Due to hardware limitations, the predicted next movement of the probe and the estimated final standard planes only accounted for the rotational component, while the translational component remained unaccounted for.} 
The performance of the imitation-based approach heavily relies on the given demonstrations. 
\secRevision{However, human US demonstrations are frequently and inherently sub-optimal}, where the sonographers often need to adjust the probe around the desired pose to finally determine the optimal view. To tackle sub-optimal demonstrations, Burke~\emph{et al.} introduced a probabilistic temporal ranking model which assumes that the images shown in the later stage are more important than the earlier images~\citep{burke2020learning}. The probabilistic ranking model can generate a large data set consisting of pair-wise images based on limited demonstrations; and then, a reward inference network was trained to assess individual B-mode images in self-supervised mode. To automatically navigate the probe to the viewpoint visualizing the mimicked tumor inside the gel phantom, an exploratory Bayesian optimization policy was employed. Nonetheless, due to safety concerns, it is \secRevision{impractical} to interact richly with patients to gain enough experience to achieve the optimal searching policy in real scenarios.

\par
The process \revision{of navigating a US probe to a proper viewpoint displaying standard planes can be seen as a series of probe motions performed in accordance} with current observations (e.g., US images, force, probe pose). Therefore, the reinforcement learning (RL) architecture has been seen as a particularly suitable solution for this type of task. Milletari~\emph{et al.} presented an initial work using a deep Q-learning (DQN) architecture to guide sonographers towards the correct sonic window for cardiac examination~\citep{milletari2019straight}. 
To avoid dynamic interaction with patients, a grid world environment was built over the chest using recorded videos to simulate acquisition environment.
\secRevision{The results demonstrated that} the DQN-based approach achieved better results ($86.1\%$ correct guidance) than a supervised approach ($77.8\%$ correct guidance) trained on the same data. A similar work also trained a DQN on a simulated 2D grid environment to navigate the probe towards the sacrum~\citep{hase2020ultrasound}. \revision{To automatically terminate the navigation process, a binary classifier (ResNet18) was employed to determine if the target object had been reached.}
Since this method only considered 3-DOF translational movements, the probe orientation is necessary to be carefully initialized.


\par
To further eliminate the requirement of manual initialization and automatically localize the paramedian sagittal oblique plane (a standard plane used in spine US examination), Li~\emph{et al.} trained a DQN to predict the potential actions in 5-DOF spaces (besides the \secRevision{translation} in the probe centerline)~\citep{li2021autonomous}. In contrast to the grid word environment, this work built a simulator using 3D US volumes that cover the target anatomy of interest. \secRevision{This simulator can generate synthetic US images based on arbitrary probe poses.} The experimental results demonstrated that the method can repeatably navigate the probe to the target standard plane with an accuracy of $4.91~mm$ (translational) and $4.65^{\circ}$ (orientational) in the intra-patient setting. Then, the authors extended the work by adding a deep learning module (VGG-16) to recognize the target standard views from real-time US images~\citep{li2021image}. Due to the US simulator, a large amount of state-action data can be obtained for training the DQN agent. In addition, to learn the policy to guide the probe to the position visualizing the kidney, Chen~\emph{et al.} used a supervised learning process to predict the next actions based on the current US image; and an actor-critic RL module was developed to improve the utilization of data and enhance the generalization~\citep{chen2021learning}. Recently, to bridge the gap between simulation and real scenarios, Bi~\emph{et al.} proposed VesNet-RL to perform US standard plane (longitudinal view) searching for vascular structures~\citep{bi2022vesnet}. To achieve high generalization capability, this study computed the binary mask of real-time B-mode images and used the background-irreverent binary masks as the input to train the RL agent. 

\par
Instead of performing validation in the simulated environment with a virtual probe, Ning~\emph{et al.} proposed a state representation model to encode the force and US images into the scene image space acquired using an RGB camera; and then an agent was trained using the proximal policy optimization (PPO) method to control the robotic manipulator to automatically perform US scans in real world~\citep{ning2021autonomic}. Similarly, Deng~\emph{et al.} employed a deep neural network to encapsulate the scanning skill (the US images, the pose/position of the probe, and the contact force) into a high-dimensional multi-modal model; then, a policy was trained based on expert demonstrations~\citep{deng2021learning}. Due to the differences between the images in the given demonstrations and real \secRevision{ones} obtained during dynamic interactions, the trained model was further improved with guided explorations carried out by human operators. However, such manual correction is very expensive during clinical examinations, and it will limit the efficiency of the RUSS. 
\final{Instead of directly learning a policy to search for standard planes, Jiang~\emph{et al.} proposed a novel machine learning framework (MI-GPSR) to understand the implicit physiological knowledge from expert demonstrations, which is implemented in a fashion of self-supervised mode using a probability ranking approach~\citep{jiang2023intelligent}. To ensure the generalization capability of the method, the authors employed the mutual information~\citep{bi2023mi} to explicitly disentangle the task-related features from the domain features. The results on three types of phantoms [gel tubular structure, chicken heart, and lamb kidney phantom (see Fig.~\ref{Fig_AI_based_RUSS})] demonstrated that MI-GPSR can properly predict the reward of individual US images from unseen demonstrations and unseen phantoms with the same anatomy~\citep{jiang2023intelligent}. Understanding and modeling the semantic reasoning and intention of expert sonographers can facilitate not only the development of autonomous intelligent RUSS but also the design of US education and training systems and advanced methods for grading and evaluating the performance of human and robotic sonography.
}

\section{Open Challenges \revision{and Future Perspectives}}~\label{sec_VII_furtu}
\par
Medical robots have gained increased attention, in particular during the COVID-19 pandemic. The role of robotics in managing public health and infectious diseases has been widely discussed among the community~\citep{hager2020role, zemmar2020rise, di2021medical, yang2020combating, gao2021progress, khamis2021robotics}. In order to apply RUSS in clinical practice, there are still many open challenges, including both technological (e.g., deep understanding of the dynamic scene, and advanced sensing technologies) and nontechnological (e.g., regulatory affairs and financing) aspects~\citep{takacs2021fasttracking, fichtinger2022image}. Here, we highlight two aspects that will widely affect the roadmap for RUSS, particularly for clinical translation and commercialization: 1) the acceptance of RUSS, and 2) the ethical and legal issues. In addition, we discussed some promising research directions to inspire the future development of RUSS.

\subsection{Acceptance by Patients and Clinicians}
\par
The RUSS are designed to help both sonographers and patients in clinical practice. Besides demonstrating comparable or even better outcomes, the acceptance for RUSS is also important. Here, we want to first make a distinction between the concepts of acceptance and trust. Trust is mostly based on how well RUSS performs in terms of technical performance, such as safety, clinical results, robustness, repeatability, and so on. Yet, effective communication, friendly interaction, and mental development would also be necessary for improving acceptance.

\par
Regarding teleoperated RUSS, Adams~\emph{et al.} indicated that all patients ($18$) were willing ($89\%$ were strongly willing and the remaining $11\%$ were willing) to have another telerobotic examination~\citep{adams2017initial}. A similar result was reported by~\citep{adams2018crossover}, where $97\%$ of $28$ patients were willing to have another teleoperation scan. \revision{However, the number of participating patients in these two studies is limited. A more comprehensive survey about the patients' acceptance of RUSS should be carried out in the future. Furthermore, it is noteworthy that the clinicians' attitudes toward RUSS are still missing.}

\par
Teleoperation systems are controlled by human operators, and there are some very successful teleoperation surgical systems, e.g., da Vinci system. This fact contributes to the positive attitude of stakeholders for teleoperated RUSS~\citep{adams2017initial, adams2018crossover}. In contrast, since autonomous RUSS are partially or fully out of the control of experts, non-negligible worries about safety arise, which stress both patients and experts during scans. Autonomous RUSS is still far from gaining widespread acceptance.
\secRevision{A standard evaluation metric considering clinical practices will help improve the trustiness of emerging autonomous medical robotics~\citep{haidegger2019autonomy}.} Nagy~\emph{et al.} defined the concept of level of Clinical Realism: 1) Training tasks with rigid phantoms; 2) Surgical tasks with simple phantoms; 3) Surgical tasks with realistic phantoms, but little or no soft-tissue interaction; 4) Surgical tasks with soft-tissue interaction; 5) Surgical tasks with soft-tissue topology changes~\citep{nagy2022performance}.

\par
To tackle the safety concern of autonomous RUSS, robotic arms are often controlled in compliant force mode, which will result in soft interaction between the probe and patients to prevent excessive contact force~\citep{osburg2022using, jiang2021autonomous_TIE_vessel, hennersperger2016towards}. A force threshold is specified as a hard limitation in the low-level controllers to completely eliminate the potential extreme situation. The RUSS will stop instantly whenever the real-time force exceeds the predetermined threshold, which was $25~N$ in~\citep{osburg2022using, jiang2021autonomous_TIE_vessel, hennersperger2016towards}. During robotic scans, two emergency buttons are often held by the clinical expert and the patient, respectively, to incorporate their observations into the safety-aware loop. Such a dedicated multi-layer safety-aware framework is beneficial for increasing the trust of clinicians and patients. By offering detailed explanations of the ongoing robotic US scans over audio and doing some straightforward interactions with patients such as ''high five", Eilers~\emph{et al.} claimed that the acceptance from patients could be enhanced~\citep{eilers2023importance}.

\par
\secRevision{
To improve the acceptance of new medical devices in clinical practices, the robotic system with a medical certification can speed up the process in both research and market-driven developments~\citep{takacs2021fasttracking}.} For example, KUKA LBR iiwa has been widely used as the key component for developing RUSS~\citep{jiang2021autonomous_TIE_vessel, osburg2022generalized}. Nevertheless, this comes with a high unit cost and may necessitate the assistance of an experienced engineer for imaging acquisition or routine system maintenance~\citep{takacs2021fasttracking}. Since the fee will be paid by the end-users, the financial issue will become a practical factor hindering the acceptance from the patients. \secRevision{Most recently, Kosa~\emph{et al.} examined the role of robotics in Intensive Care Medicine and their acceptability to patients and caregivers~\citep{kosa2023robots}. They concluded that it is still immature to use robots directly handling patients, and close collaborations between \final{roboticists} and clinicians are required to advance robotics to benefit the ICU.}

\subsection{Ethical and Legal Issues}
\par
The ethical and legal issues regarding medical robotics are still not clearly defined, particularly for autonomous systems. The distribution of responsibility between experts and RUSS (or other surgical robotic systems) remains unclear. Clinical translation will also need regulatory acceptance.

\par
In order to properly tackle the ethical, regulatory, and legal issues for RUSS, Yang~\emph{et al.} divided surgical robots into six subgroups in terms of autonomy levels: no autonomy, robot assistance, task autonomy, conditional autonomy, high autonomy, and full autonomy~\citep{yang2017medical}. \secRevision{To further improve the concept of level of autonomy, Haidegger defined the term ``situation awareness" as the operator’s perception, comprehension, and prediction of a robot’s behavior in its environment~\citep{haidegger2019autonomy}. Then, ``situation awareness" is used to distinguish the required level of human supervision.} 
Up to the time of writing this article, commercial surgical robots are still solidly resting at Level-0, while a very large number of high-autonomy surgical robotic systems are waiting for clinical translation~\citep{attanasio2021autonomy}. Since commercial surgical robots are dominated by a few disproportionately large companies; thereby they have no rush in disrupting the status quo~\citep{attanasio2021autonomy}. 
\revision{
Ethical and legal regulations are critical for clinical translation and further commercialization. The need for such a regulation has been highlighted by various senior researchers in multiple impactful publications recently~\citep{fichtinger2022image, khamis2019ai, haidegger2019autonomy, prestes2021first, haidegger2022robot}.}

\par
To establish such regulations for medical robots, O'Sullivan~\emph{et al.} defined three different responsibilities: (1) accountability: the capacity of a system to give an explanation for its actions; (2) liability: the legal liability for potential damages caused by a robot; and (3) culpability: whom and how to implement punishment~\citep{o2019legal}. In addition, Vayena~\emph{et al.} discussed ethical and legal issues for digital health in terms of privacy and security, trust, and accountability~\citep{vayena2018digital}. As a large amount of data is often necessary for analysis, protecting privacy is undoubtedly important for avoiding misuse. Public trust is of paramount importance. Vayena~\emph{et al.} considered that the creation of a culture of trust will enable all stakeholders to benefit from the development of digital health~\citep{vayena2018digital}. Similarly, Yang~\emph{et al.} summarised five increasingly pressing topics in terms of ethics for robotics and AI~\citep{yang2018grand}. Besides the aforementioned terms like responsibility, this works further emphasized some societal issues such as potential influence on employment and human freedom. Due to the quick evolution of the area of medical robotics, a proper and comprehensive regulatory system will boost a prosperous market and gradually benefit all stakeholders.

\par
\revision{
To deal with the unsolved issues regarding the safety, transparency, and trustworthiness of modern medical devices with a certain level of autonomy, the two leading Standard Development Organizations International Organization for Standardization (ISO) and the International Electrotechnical Commission (IEC) created the first joint standardization document (IEC/TR 60601-4-1) regarding autonomy for technical developers~\citep{haidegger2019autonomy}. Recently, Prestes~\emph{et al.} established the first global ontological standard for AI and robotics: IEEE 7007—Ontological Standard for Ethically Driven Robotics and Automation Systems~\citep{prestes2021first}. For an in-depth review of the ongoing initiatives regarding regulations, we highly recommend that readers refer to these two articles~\citep{haidegger2019autonomy, prestes2021first}. 
}

\subsection{\revision{Future Perspectives}}
\revision{
In addition to challenges, there are also numerous opportunities in the field of RUSS, particularly in light of the boom in both fundamental sensor development and advanced AI research. This survey will elaborate on future perspectives from these two aspects. By providing an understanding of the state of the art, we hope it can stimulate a number of exciting ideas. To clarify, the opportunities extend far beyond what are described below.}

\subsubsection{Fundamental Sensing Systems}
\par
Sensors are essential components of all intelligent systems. Generally, the development of new sensors has a substantial effect on existing systems in numerous ways. To achieve the ultimate goal of an autonomous RUSS, it is necessary to integrate multiple sensing systems mimicking the sophisticated human sensing system. By developing efficient data fusion techniques, redundancy, and multi-modality data would aid in achieving robust and reliable perception results. This applies not only to RUSS but to \secRevision{a vast array of autonomous systems.}

\par
Most recently, the novel concept and development of US patches have become attractive. Due to the advantages of \secRevision{small} size, stretchable probability, and no need for \secRevision{US} gel, it is very desired for continuous healthcare monitoring. The traditional US probes are rigid and bulky, making them unsuitable for imaging through nonplanar surfaces. To address this challenge, Hu~\emph{et al.} proposed a stretchable US probe that can conform to and detect nonplanar complex  surfaces~\citep{hu2018stretchable}. This soft probe consisted of a $10\times 10$ array of piezoelectric transducers covered by compliant silicone elastomers, and the results demonstrated that it could be stretched more than $50\%$. Similarly, Wang~\emph{et al.} developed and tested a skin-conformal ultrasonic phased array to monitor the physiological signals from tissues up to $14~cm$~\citep{wang2021continuous}. To tackle the practical issue that the image quality is highly affected by US gels, Wang~\emph{et al.} designed a bioadhesive US device consisting of a thin and rigid US probe robustly adhered to the skin via a couplant made of a soft, tough, antidehydrating, and bioadhesive hydrogel-elastomer hybrid~\citep{wang2022bioadhesive}. Based on this device, continuous imaging of internal tissues over days becomes feasible. Most recently, Hu~\emph{et al.} demonstrate a wearable cardiac US imager providing direct cardiac function assessment~\citep{hu2023wearable}. Such fundamental changes in US probe would open numerous opportunities for revolutionizing the techniques of robot-assisted US imaging. 


\subsubsection{Advanced AI-based RUSS}
\par
We consider the AI-based RUSS would be another promising direction, where the core task is to improve the intelligence of RUSS. To this end, the research community needs first to improve the computer's understanding of dynamic environments through multi-modality signals. Only when the system owns precise perception abilities, we can further expect and explore the way to make proper decisions autonomously.
Several studies have demonstrated that AI-based approaches outperformed conventional \secRevision{image processing} methods~\cite{bass2021ultrasound, hesamian2019deep, shan2021ultrasound}.
Benefiting from the accurate segmentation of target objects (e.g., blood vessels), \secRevision{precise state representations will further facilitate the development of autonomous scanning~\cite{jiang2021autonomous_TIE_vessel, huang2021towards} or autonomous exploration of standard US planes~\citep{bi2022vesnet, li2021autonomous}.}

\par
\revision{
In addition, advanced learning-based frameworks have the potential to be used to transfer senior sonographers' physiological knowledge and experience to novices. Recent studies in the direction of learning from demonstrations~\citep{burke2020learning} implicitly result in an attractive and influential new research topic on recovery of ``language of sonography". Hands-on experience is very important and necessary for sonographers. Senior sonographers who can perform flawless US scans are still unable to directly parameterize and intuitively describe the acquisition requirements. However, US examinations are carried out based on their understanding of high-level physiological knowledge. Such knowledge is common among sonographers, although their comprehension may vary slightly due to experience. The concept of recovery of ``language of sonography" refers to the underlying understanding of high-level anatomical knowledge. We believe that efforts to \secRevision{retract} the ``language of sonography" from intuitive demonstrations with multiple signals, such as US images, RGB-D images, force information, probe movement, gaze information, etc., are as valuable and essential as the progress made in robotic sonography itself~\citep{droste2020automatic, men2022multimodal, alsharid2022gaze}.}

\section{Discussion}
\par
\secRevision{
Robotic technologies have demonstrated promising potential to extend the use of US imaging in the healthcare industry, such as remote examinations, and accurate and quantitative control of acquisition parameters.}
Compared with conventional US examinations, although current RUSS cannot yet show superiority in terms of improving clinical outputs, a number of benefits have been demonstrated. \secRevision{From the perspective of patients}, the waiting time for the healthcare intervention was significantly reduced from $144$ to $26.5$ days~\citep{boman2014robot} and their cost was reduced as well~\citep{sekar2007telecardiology}. As for sonographers, robots bring dexterity as well as reduce work-related musculoskeletal disorders~\citep{brown2003work, fang2017force}. 
Additionally, RUSS has the potential to make a significant contribution in a variety of clinical scenarios, including performing trauma examinations in pre-hospital settings~\citep{mylonas2013autonomous}, freeing up a clinician's hand during the intervention~\citep{esteban2018robotic}, and performing routine PAD screening or monitoring without radiation~\citep{jiang2021autonomous_TIE_vessel}. When it comes to trauma scans, it is vital to spot life-threatening intracavitary hemorrhage as soon as possible because this will enable doctors to make prompt treatment decisions to save lives in emergency scenarios. RUSS could be used for reliable and accurate trauma scan identification in pre-hospital settings by fusing precise sensing devices with a cutting-edge learning-based semantic segmentation framework.

\par
Continuing the current progress on RUSS requires a deep understanding of how its embedded technologies add value to healthcare practices. Intelligent robotic imaging systems could provide different benefits. On one hand, they can democratize the healthcare by making US examination available at locations in which patient populations do not currently have access to expert sonographers. On the other hand, to maximize the added value of RUSS, it is important to also focus on enabling new types of interventions or new procedures that are impractical or impossible based on traditional US examination, e.g., 3D or 4D visualization of scanned anatomy compensating or embedding physical breathing and heartbeat. Although there is not yet any fully autonomous system for US examinations, autonomy is one of the main objectives of the scientific community. \revision{Similar to surgical robotics, autonomous RUSS will be more challenging to commercialize~\citep{dupont2021decade}, however, due to its nature of offering images and visualization rather than decision making, cutting, and suturing tissues, we believe autonomous RUSS is easier to be certified and productized than autonomous surgical robotic solutions. On the other hand, compared to robotic X-ray and nuclear imaging, RUSS may be harder to certify because it requires direct interaction with patients. Researchers, therefore, need to continue their studies to guarantee the trust in and acceptance of autonomous RUSS by both doctors and patients.}

\par
The reported results on current autonomous RUSS are still far from maturity and do not perform as well as or outperform clinicians.
Most existing research makes simplifying assumptions and often uses artificial setups for their validation. For example, most US servoing approaches (Section~\ref{sec_V_C_Adv_servo}) are validated on phantoms or using simulation rather than on human subjects, and the existing motion and deformation compensation approaches may not perform as well on patients within the complex and dynamic clinical setups.

\par
\begin{enumerate}
  \item Could advanced machine learning allow us to learn the ``language of sonography" by observing expert sonographers?
  \item Could our RUSS systems understand the physics of imaging and its interaction with dynamic patient physiology?
  \item Could RUSS allow optimizing B-Mode, 3D and 4D image acquisition?
  \item Could advanced sensing and intelligent control allow for guaranteeing reproducibility and safety of scanning procedures?
  \item Could multimodal imaging and pretraining allow RUSS systems to observe and understand the specific anatomy and physiology of each patient?
  \item Could explainable AI enable RUSS systems to report and justify their actions and decisions to physicians?
  \item Could user-centric RUSS design allow smooth and friendly communication between sonographer robots, physician colleagues, and patients?
\end{enumerate}
Answering each of these exciting and essential questions requires large multi-disciplinary scientific and engineering communities to gather, communicate and collaborate. \revision{The current review paper hopes to play a small role in gathering and highlighting some of the requirements and opening the path for the community to study and analyze the next crucial steps to take.}

\section{Conclusion}
\par
This survey has provided a brief picture of the rapidly evolving field of robot-assisted US imaging systems. Starting from the technical developments and clinical translations of various teleoperation systems in the first decade of the new millennium, in Section~\ref{sec_III_tele}, the article summarizes the path the community took to get to its recent research focus on autonomous RUSS, in particular after the booming of machine learning and artificial intelligence throughout the last decade. 
It is challenging to develop intelligent RUSS solutions, which require a number of advanced capabilities to understand dynamic environments, physics of US imaging, human anatomy and physiology, and thereby to tackle complex cases of diagnostic and interventional imaging. 
To date, there are no such systems available. This paper aims at reviewing the state of the art and discussing the paths the community has taken or needs to take in the future. 

\par
The survey shows that the recent progress has demonstrated that RUSS may be able to improve image acquisition and 3D visualization, also taking motion and deformation into account, real-time geometrical (including volumetric) measurements, and in particular their reproducibility. The US handling habits vary among expert sonographers, and cannot be well described using \revision{handcrafted} features. We believe that in \revision{the} near future, the development of advanced machine learning will allow for figuring out the underlying ``language of sonography" based on expert demonstrations. This can not only allow for autonomous intelligent RUSS development but also for designing US education and training systems, and advanced methodologies for grading and evaluating the performance of human and robotic US examinations. In view of its speed of progress, RUSS has the potential to revolutionize not only the US-based medical interventions themselves but also clinical screening, diagnosis, and robotic-assisted surgery.











\section*{Declaration of Competing Interest}
\label{competinginterest}

The authors report no conflicts of interest.

\section*{Acknowledgments}
\label{acknowledgments}
The authors would like to acknowledge the Editors and anonymous reviewers for their time, and \secRevision{implicit contributions} to the improvement of the article's thoroughness, readability, and clarity.


\bibliographystyle{model2-names.bst}\biboptions{authoryear}
\bibliography{references}



\end{document}